\def\year{2016}\relax
\documentclass[letterpaper]{article}
\usepackage{aaai16}
\usepackage{times}
\usepackage{helvet}
\usepackage{courier}

\usepackage{times}
\usepackage{url}
\usepackage{latexsym}
\usepackage{graphicx}
\usepackage{color}
\usepackage{CJK}
\usepackage{fancyhdr}
\usepackage{multirow}
\usepackage{amsfonts}
\usepackage{amsmath}
\usepackage{hyperref}
\usepackage{algorithm}
\usepackage{algpseudocode}

\begin{document}
%
\title{Probabilistic Reasoning via Deep Learning: Neural Association Models}
\author{
Quan Liu$^\dagger$, Hui Jiang$^\ddagger$, Andrew Evdokimov$^\ddagger$, Zhen-Hua Ling$^\dagger$, Xiaodan Zhu$^\ell$,  Si Wei$^\S$, Yu Hu$^\dagger$$^\S$\\
$^\dagger$ National Engineering Laboratory for Speech and Language Information Processing \\
University of Science and Technology of China, Hefei, Anhui, China \\
$^\ddagger$ Department of Electrical Engineering and Computer Science, York University, Canada\\
$^\ell$ National Research Council Canada, Ottawa, Canada\\
$\S$ iFLYTEK Research, Hefei, China \\
\em emails: quanliu@mail.ustc.edu.cn, hj@cse.yorku.ca, ae2718@cse.yorku.ca, zhling@ustc.edu.cn \\ 
\em xiaodan@cse.yorku.ca, siwei@iflytek.com, yuhu@iflytek.com
}
\maketitle

\begin{abstract}
In this paper, we propose a new deep learning approach, called neural association model (NAM), for probabilistic reasoning in artificial intelligence.
We propose to use neural networks to model association between any two events in a domain.
Neural networks take one event as input and compute a conditional probability of the other event to model how likely these two events are to be associated.
The actual meaning of the conditional probabilities varies between applications and depends on how the models are trained.
In this work, as two case studies, we have investigated two NAM structures, namely deep neural networks (DNN) and relation-modulated neural nets (RMNN), on several probabilistic reasoning tasks in AI, including recognizing textual entailment, triple classification in multi-relational knowledge bases and commonsense reasoning.
Experimental results on several popular datasets derived from WordNet, FreeBase and ConceptNet have all demonstrated that both DNNs and RMNNs perform equally well and they can significantly outperform the conventional methods available for these reasoning tasks.
Moreover, compared with DNNs, RMNNs are superior in knowledge transfer, where a pre-trained model can be quickly extended to an unseen relation after observing only a few training samples.
To further prove the effectiveness of the proposed models, in this work, we have applied NAMs to solving challenging \textit{Winograd Schema} (WS) problems. 
Experiments conducted on a set of WS problems prove that the proposed models have the potential for commonsense reasoning.
\end{abstract}

\section{Introduction}
\label{sec:intro}
Reasoning is an important topic in artificial intelligence (AI), which has attracted considerable attention and research effort in the past few decades \cite{mccarthy1986applications,minsky1988society,mueller2014commonsense}.
Besides the traditional logic reasoning, probabilistic reasoning has been studied as another typical genre in order to handle knowledge uncertainty in reasoning based on probability theory \cite{pearl1988probabilistic,neapolitan2012probabilistic}.
The probabilistic reasoning can be used to predict conditional probability $\Pr(E_{2}|E_{1})$ of one event $E_{2}$ given another event $E_{1}$.
State-of-the-art methods for probabilistic reasoning include
Bayesian Networks \cite{jensen1996introduction},
Markov Logic Networks \cite{richardson2006markov}
and other graphical models \cite{koller2009probabilistic}.
Taking Bayesian networks as an example, the conditional probabilities between two associated events are calculated as posterior probabilities according to Bayes theorem, 
with all possible events being modeled by a pre-defined graph structure.
However, these methods quickly become intractable for most practical tasks where the number of all possible events is usually very large.

In recent years, distributed representations that map discrete language units into continuous vector space have gained significant popularity along with the development of neural networks \cite{bengio2003neural,collobert2011natural,mikolov2013efficient}.
The main benefit of embedding in continuous space is its smoothness property, which helps to capture the semantic relatedness between discrete events, potentially generalizable to unseen events.
Similar ideas, such as knowledge graph embedding, have been proposed to represent knowledge bases (KB) in low-dimensional continuous space \cite{bordes2013translating,socher2013reasoning,wang2014knowledge,nickel2015review}.
Using the smoothed KB representation, it is possible to reason over the relations among various entities.
However, human-like reasoning remains as an extremely challenging problem partially because it requires the effective encoding of world knowledge using powerful models.
Most of the existing KBs are quite sparse and even recently created large-scale KBs, such as YAGO, NELL and Freebase, can only capture a fraction of world knowledge.
In order to take advantage of these sparse knowledge bases, the state-of-the-art approaches for knowledge graph embedding usually adopt simple linear models, such as RESCAL \cite{nickel2012factorizing}, TransE \cite{bordes2013translating} and Neural Tensor Networks \cite{socher2013reasoning,bowman2013can}. 

Although deep learning techniques achieve great progresses in many domains, e.g. speech and image \cite{lecun2015deep}, the progress in commonsense reasoning seems to be slow.
In this paper, we propose to use deep neural networks, called {\em neural association model (NAM)}, for commonsense reasoning.
Different from the existing linear models, the proposed NAM model uses multi-layer nonlinear activations in deep neural nets to model the association conditional probabilities between any two possible events.
In the proposed NAM framework, all symbolic events are represented in low-dimensional continuous space
and there is no need to explicitly specify any dependency structure among events as required in Bayesian networks.
Deep neural networks are used to model the association between any two events, taking one event as input to  compute a conditional probability of another event.
The computed conditional probability for association may be generalized to model various reasoning problems, such as entailment inference, relational learning, causation modelling and so on.
In this work, we study two model structures for NAM.
The first model is a standard deep neural networks (DNN) and the second model uses a special structure called relation modulated neural nets (RMNN).
Experiments on several probabilistic reasoning tasks, including recognizing textual entailment, triple classification in multi-relational KBs and commonsense reasoning,
have demonstrated that both DNNs and RMNNs can outperform other conventional methods.
Moreover, the RMNN model is shown to be effective in knowledge transfer learning, where
a pre-trained model can be quickly extended to a new relation after observing only a few training samples.

Furthermore, we also apply the proposed NAM models to more challenging commonsense reasoning problems, i.e., the recently proposed \textit{Winograd Schemas} (WS) \cite{levesque2011winograd}.
The WS problems has been viewed as an alternative to the Turing Test \cite{turing1950computing}.
To support the model training for NAM, we propose a straightforward method to collect associated cause-effect pairs from large unstructured texts.
The pair extraction procedure starts from constructing a vocabulary with thousands of common verbs and adjectives. 
Based on the extracted pairs, this paper extends the NAM models to solve the \textit{Winograd Schema} problems and achieves a 61\% accuracy on a set of cause-effect examples. 
Undoubtedly, to realize commonsense reasoning, there is still much work be done and many problems to be solved. 
Detailed discussions would be given at the end of this paper.

\section{Motivation: Association between Events}
\label{sec:motivation}
This paper aims to model the association relationships between events using neural network methods.
To make clear our main work, we will first describe the characteristics of events and all the possible association relationships between events.
Based on the analysis of event association, we present the motivation for the proposed neural association models.
In commonsense reasoning, the main characteristics of events are the following:
\begin{itemize}
\item \textbf{Massive}: In most natural situations, the number of events is massive, which means that the association space we will model is very large.
\item \textbf{Sparse}: All the events occur in our dialy life are very sparse. It is a very challenging task to ideally capture the similarities between all those different events.
\end{itemize}

\begin{figure}[htb]
  \centering
  \includegraphics[width=7cm]{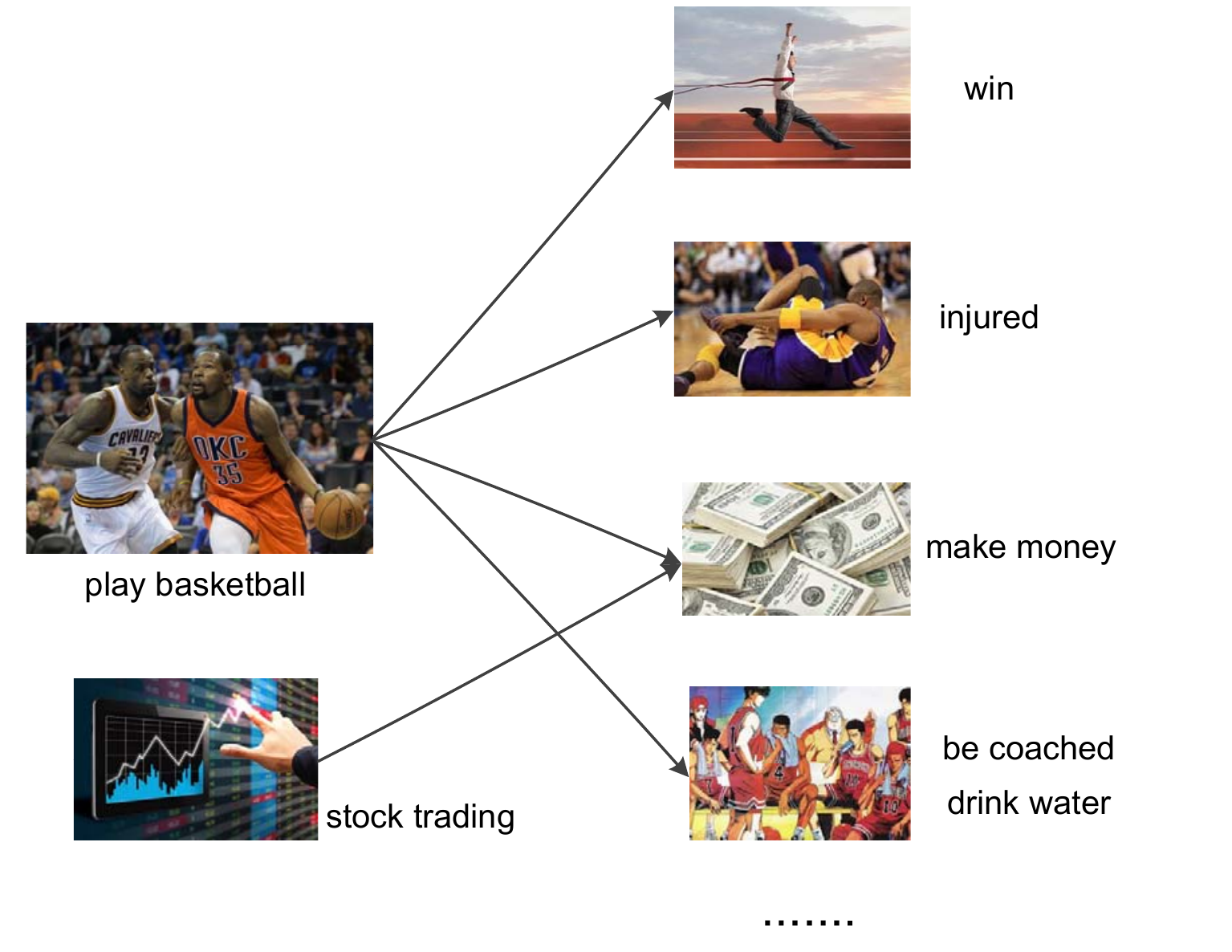}\\
  \caption{Example of association between events.}
  \label{fig:asso-example}
\end{figure}

At the same time, association between events appears everywhere. Consider a single event \textit{play basketball} for example, shown in Figure \ref{fig:asso-example}.
This single event would associate with many other events. 
A person who plays basketball would win a game. Meanwhile, he would be injured in some cases.
The person could make money by playing basketball as well.
Moreover, we know that a person who plays basketball should be coached during a regular game.
Those are all typical associations between events.
However, we need to recognize that the task of modeling event \textit{association} is not identical to performing \textit{classification}.
In classification, we typically map an event from its feature space into one of pre-defined finite categories or classes. 
In event association, we need to compute the association probability between two arbitrary events, each of which may be a sample from a possibly infinite set. 
The mapping relationships in event association would be \textit{many-to-many}; e.g., not only playing basketball could support us to make money, someone who makes stock trading could make money as well.
More specifically, the association relationships between events include cause-effect, spatial, temporal and so on. 
This paper treats them as a general relation considering the sparseness of useful KBs.

In this paper, we believe that modeling the the association relationships between events is a fundamental work for commonsense reasoning. 
If we could model the event associations very well, we may have the ability to solve many commonsense reasoning problems.
Considering the main characteristics of \textit{discrete event} and \textit{event association}, two reasons are given for describing our motivation. 
\begin{itemize}
\item The advantage of distributed representation methods: representing discrete events into continuous vector space provides a good way to capture the similarities between discrete events.
\item The advantage of neural network methods: neural networks could perform universal approximation while linear models cannot easily do this \cite{hornik1990universal}. 
\end{itemize}

At the same time, this paper takes into account that both distributed representation and neural network methods are data-hungry.
In Artificial Intelligence (AI) research, mining large sizes of useful data (or knowledge) for model learning is always challenging.
In the following section, this paper presents a preliminary work on data collection and the corresponding experiments we have made for solving commonsense reasoning problems.
\section{Neural Association Models (NAM)}
\label{sec:main}
In this paper, we propose to use a nonlinear model, namely neural association model, for probabilistic reasoning.
Our main goal is to use neural nets to model the association probability for any two events $E_1$ and $E_2$ in a domain, i.e., $\Pr(E_{2}|E_{1})$ of $E_{2}$ conditioning on $E_{1}$.
All possible events in the domain are projected into continuous space without specifying any explicit dependency structure among them.
In the following, we first introduce neural association models (NAM) as a general modeling framework for probabilistic reasoning.
Next, we describe two particular NAM structures for modeling the typical multi-relational data.

\subsection{NAM in general}
\label{ssec:nam-general}
\begin{figure}[htb]
  \centering
  \includegraphics[width=8.5cm]{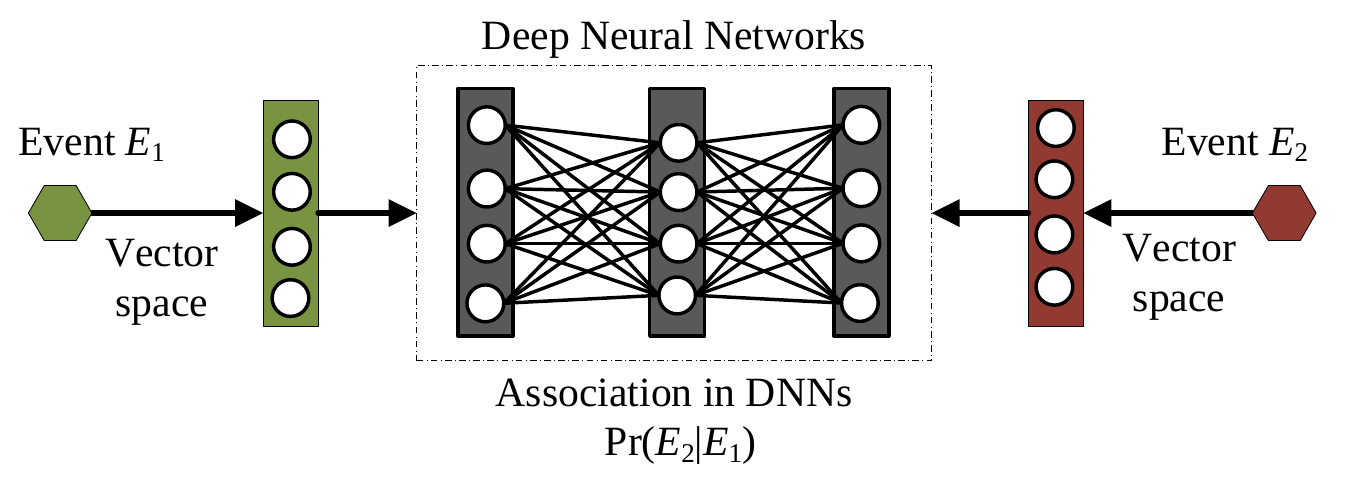}\\
  \caption{The NAM framework in general.}
  \label{fig:NAM-general}
\end{figure}
Figure \ref{fig:NAM-general} shows the general framework of NAM for associating two events, $E_{1}$ and $E_{2}$.
In the general NAM framework, the events are first projected into a low-dimension continuous space.
Deep neural networks with multi-layer nonlinearity are used to model how likely these two events are to be associated.
Neural networks take the embedding of one event $E_{1}$ (antecedent) as input and compute a conditional probability $\Pr(E_{2}|E_{1})$ of the other event $E_{2}$ (consequent).
If the event $E_2$ is binary (true or false), the NAM models may use a \textit{sigmoid} node to compute $\Pr(E_{2}|E_{1})$. If $E_2$ takes multiple mutually exclusive values, we use a few \textit{softmax} nodes for $\Pr(E_{2}|E_{1})$, where it may need to use multiple embeddings for $E_2$ (one per value).
NAMs do not explicitly specify how different events $E_2$ are actually related;
they may be mutually exclusive, contained, intersected.
NAMs are only used to separately compute conditional probabilities, $\Pr(E_{2}|E_{1})$,  for each pair of  events, $E_1$ and $E_2$, in a task.
The actual physical meaning of the conditional probabilities $\Pr(E_{2}|E_{1})$ varies between applications and depends on how the models are trained.
Table \ref{tab:app-physical} lists a few possible applications.
\begin{table}[htb]
\centering
\begin{tabular}{c|c|c}
  \hline
  Application & $E_{1}$ & $E_{2}$ \\\hline
  language modeling & $h$ & $w$ \\
  causal reasoning & \textit{cause} & \textit{effect} \\
  knowledge triple classification & $\{e_{i},r_{k}\}$ & $e_{j}$ \\
  lexical entailment & $W_{1}$ & $W_{2}$ \\
  textual entailment & $D_{1}$ & $D_{2} $\\
  \hline
\end{tabular}
\caption{Some applications for NAMs.}
\label{tab:app-physical}
\end{table}

In language modeling, the antecedent event is the representation of historical context, $h$, and the consequent event is the next word $w$ that takes one out of K values.
In causal reasoning, $E_{1}$ and $E_{2}$ represent \textit{cause} and \textit{effect} respectively. For example, we have $E_1$ = \textit{``eating cheesy cakes''} and $E_2$ = \textit{``being happy''}, where $\Pr(E_2|E_1)$ indicates how likely it is that $E_1$ may cause the binary (true or false) event $E_2$. In the same model, we may add more nodes to model different effects from the same $E_1$, e.g.,  $E'_2$ = \textit{``growing fat''}. Moreover, we may add 5 softmax nodes to model a multi-valued event, e.g., $E''_2$ = \textit{``happiness''  (scale from 1 to 5)}.
Similarly, for knowledge triple classification of multi-relation data, given one triple $(e_{i}, r_{k}, e_{j})$,  $E_{1}$ consists of the head entity (\textit{subject}) $e_{i}$ and relation (\textit{predicate}) $r_{k}$, and  $E_{2}$ is a binary event indicating whether the tail entity (\textit{object}) $e_{j}$ is true or false.
Finally, in the applications of recognizing lexical or textual entailment, $E_{1}$ and $E_{2}$ may be defined as \textit{premise} and \textit{hypothesis}.
More generally, NAMs can be used to model an infinite number of events $E_2$, where each point in a continuous space represents a possible event. In this work, for simplicity, we only consider NAMs for a finite number of binary events $E_2$  but the formulation can be easily extended to more general cases.

Compared with traditional methods, like Bayesian networks, NAMs employ neural nets as a universal approximator  to directly model individual pairwise event association probabilities without relying on explicit dependency structure. Therefore, NAMs can be end-to-end learned purely from training samples without strong human prior knowledge, and are potentially more scalable to real-world tasks.


\subsubsection{Learning NAMs}
\label{sssec:main-objective}

Assume we have a set of $N_{d}$ observed examples (event pairs $\{ E_1, E_2 \}$), $\mathcal{D}$, each of which is denoted as $x_{n}$.
This training set normally includes both positive and negative samples.
We denote all positive samples ($E_2=true$) as $\mathcal{D}^{+}$ and
all negative samples ($E_2=false$) as $\mathcal{D}^{-}$.
Under the same independence assumption as in statistical relational learning (SRL) \cite{getoor2007introduction,nickel2015review}, the log likelihood function of a NAM model can be expressed as follows:
\begin{equation} \label{eq:NAM-OBJ}
\begin{aligned}
  \mathcal{L}(\mathbf{\Theta}) =
  & \sum_{x_{n}^{+} \in \mathcal{D}^{+}} \ln f(x_{n}^{+}; \mathbf{\Theta})
   + \sum_{x_{n}^{-} \in \mathcal{D}^{-}} \ln (1-f(x_{n}^{-}; \mathbf{\Theta})) \\
\end{aligned}
\end{equation}
where $f(x_{n}; \mathbf{\Theta})$ denotes a \textit{logistic} score function derived by the NAM for each $x_{n}$, which numerically computes the conditional probability $\Pr(E_{2}|E_{1})$. More details on $f(\cdot)$ will be given later in the paper.  Stochastic gradient descent (SGD) methods may be used to maximize the above likelihood function, leading to a maximum likelihood estimation (MLE) for NAMs.


In the following, as two case studies, we consider two NAM structures with a finite number of output nodes to model $\Pr(E_2|E_1)$ for any pair of events, where we have only a finite number of $E_2$ and each $E_2$ is binary.
%
The first model is a typical DNN that associates antecedent event ($E_1$) at input and consequent event ($E_2$) at output. We then present another model structure, called relation-modulated neural nets,
which is more suitable for multi-relational data.

\subsection{DNN for NAMs}
\label{ssec:nam-DNNs}
\begin{figure}[htb]
  \centering
  \includegraphics[width=7cm]{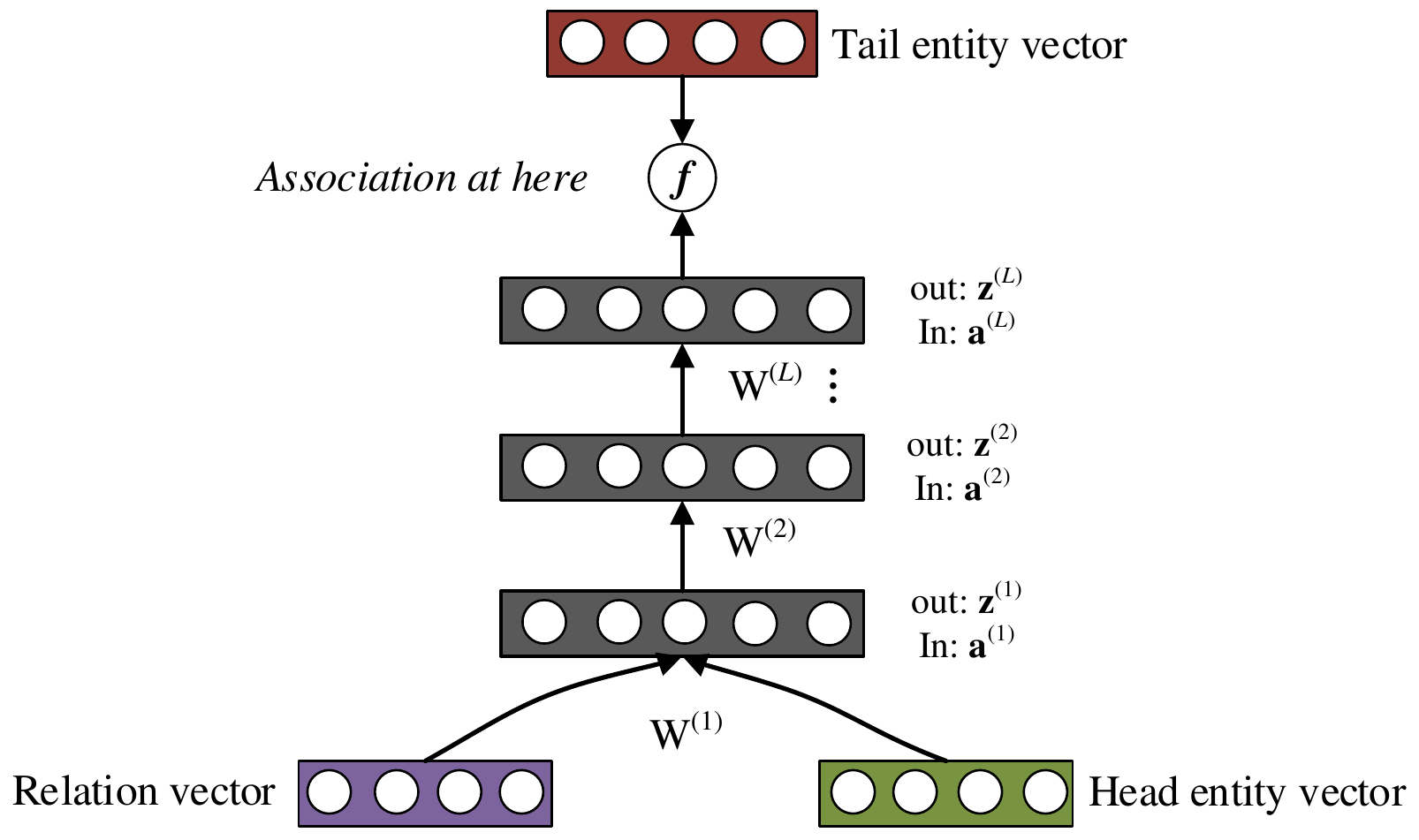}\\
  \caption{The DNN structure for NAMs.}
  \label{fig:DNNs-NAM}
\end{figure}

The first NAM structure is a traditional DNN as shown in Figure \ref{fig:DNNs-NAM}.
Here we use multi-relational data in KB for illustration.
Given a KB triple $x_{n}=(e_{i}, r_{k}, e_{j})$ and its corresponding label $y_{n}$ (true or false),
we cast $E_{1}=(e_{i},r_{k})$ and $E_{2}=e_{j}$ to compute $\Pr(E_2|E_1)$ as follows.

Firstly, we represent head entity phrase $e_{i}$ and tail entity phrase $e_{j}$ by two embedding vectors $\mathbf{v}^{(\mathrm{1})}_{i} (\in \mathbf{V}^{\mathrm{(1)}})$ and $\mathbf{v}^{(\mathrm{2})}_{j} (\in \mathbf{V}^{\mathrm{(2)}})$.
Similarly, relation $r_{k}$ is also represented by a low-dimensional vector $\mathbf{c}_{k} \in \mathbf{C}$, which we call a \textit{relation code} hereafter.
Secondly, we combine the embeddings of the head entity $e_{i}$ and the relation $r_{k}$ to feed into an $(L + 1)$-layer DNN as input.
The DNN consists of $L$  rectified linear (ReLU) hidden layers \cite{nair2010rectified}.
The input is $\mathbf{z}^{(0)} = [\mathbf{v}^{(\mathrm{1})}_{i}, \mathbf{c}_{k}]$.
During the feedforward process, we have
\begin{equation}\label{eq-layer-mul}
  \mathbf{a}^{(\ell)} = \mathbf{W}^{(\ell)} \mathbf{z}^{(\ell-1)} + \mathbf{b}^{\ell} \quad (\ell = 1, \cdots, L)
\end{equation}
\begin{equation}\label{eq-layer-act}
  \mathbf{z}^{(\ell)} = h \left( \mathbf{a}^{(\ell)} \right) = \max \left(0, \mathbf{a}^{(\ell)}\right) \quad
  (\ell = 1, \cdots, L)
\end{equation}
where $\mathbf{W}^{(\ell)}$ and $\mathbf{b}^{\ell}$ represent the weight matrix and bias for layer $\ell$ respectively.
%

Finally, we propose to calculate a sigmoid score for each triple $x_{n}=(e_{i}, r_{k}, e_{j})$ as the association probability using the last hidden layer's output and the tail entity vector $\mathbf{v}^{(\mathrm{2})}_{j}$:
%
\begin{equation}\label{eq:f-ijk}
  f(x_{n};\mathbf{\Theta}) = \sigma \left( \mathbf{z}^{(L)} \cdot \mathbf{v}^{(\mathrm{2})}_{j} \right)
\end{equation}
where $\sigma(\cdot)$ is the \textit{sigmoid} function, i.e., $\sigma(x) = 1/(1+e^{-x})$.

All network parameters of this NAM structure, represented as $\mathbf{\Theta}=\{\mathbf{W}, \mathbf{V}^{\mathrm{(1)}}, \mathbf{V}^{\mathrm{(2)}}, \mathbf{C}\}$, may be jointly learned by maximizing the likelihood function in eq. (\ref{eq:NAM-OBJ}).

\subsection{Relation-modulated Neural Networks (RMNN)}
\label{ssec:nam-RMNNs}

\begin{figure}[htb]
  \centering
  \includegraphics[width=7.25cm]{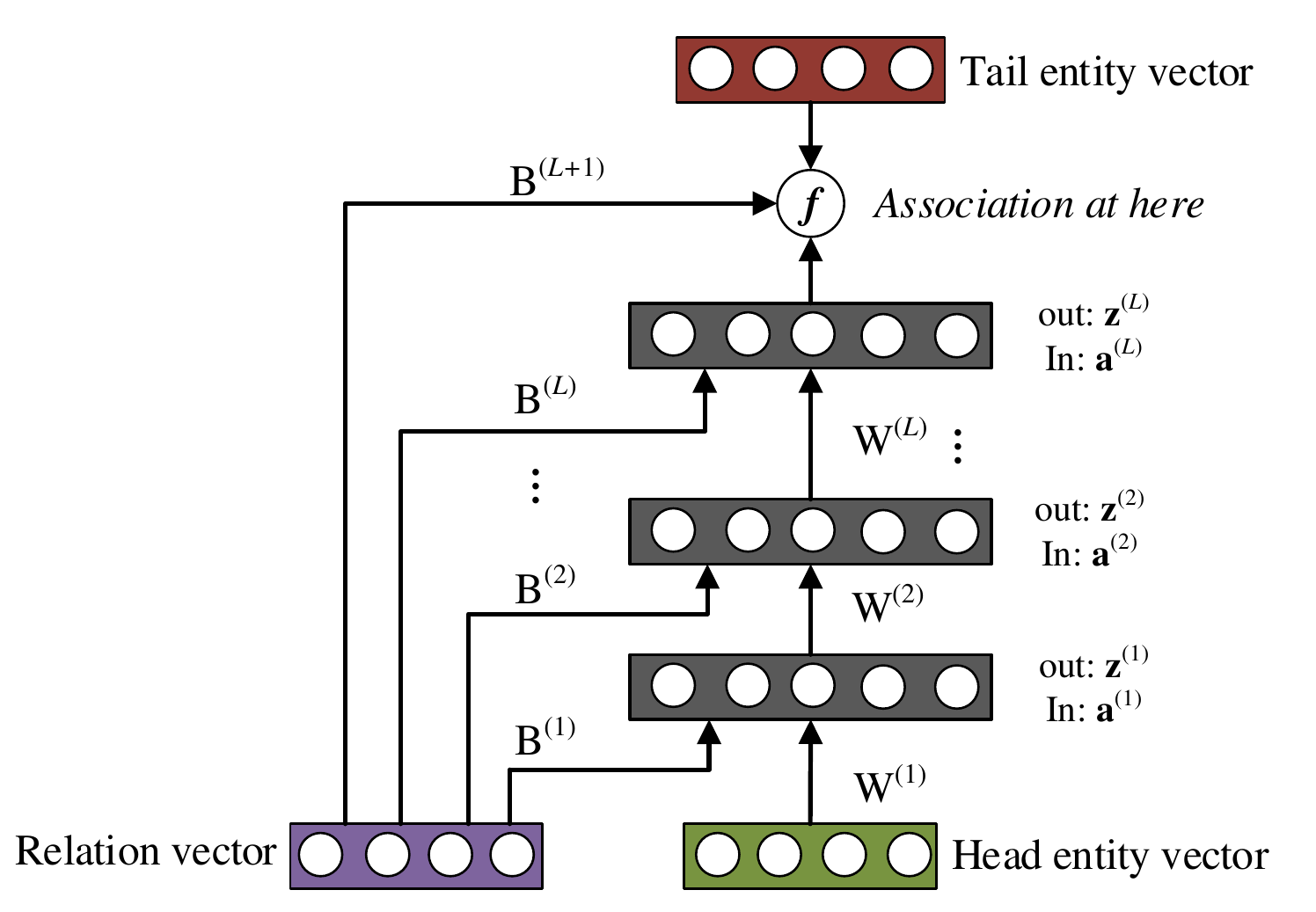}\\
  \caption{The relation-modulated neural networks (RMNN).}
  \label{fig:RMNNs}
\end{figure}

Particularly for multi-relation data, following the idea in \cite{xue2014fast}, we propose to use the so-called relation-modulated neural nets (RMNN),
as shown in Figure \ref{fig:RMNNs}.

The RMNN uses the same operations as DNNs to project all entities and relations
into low-dimensional continuous space.
As shown in Figure \ref{fig:RMNNs}, we connect the knowledge-specific relation code $\mathbf{c}^{(k)}$ to all hidden layers in the network. As shown later, this structure is superior in knowledge transfer learning tasks.
Therefore, for each layer of RMNNs, instead of using eq.(\ref{eq-layer-mul}), its linear activation signal is computed from the previous layer $\mathbf{z}^{(\ell-1)}$ and the relation code $\mathbf{c}^{(k)}$ as follows:
\begin{equation}\label{eq-layer-mul-code}
  \begin{aligned}
  \mathbf{a}^{(\ell)} & = \mathbf{W}^{(\ell)} \mathbf{z}^{(\ell-1)} + \mathbf{B}^{(\ell)} \mathbf{c}^{(k)}, \quad (\ell = 1 \cdots L) \\
  \end{aligned}
\end{equation}
where $\mathbf{W}^{(\ell)}$ and $\mathbf{B}^{\ell}$ represent the normal weight matrix and the relation-specific weight matrix for layer $\ell$.
At the topmost layer, we calculate the final score for each triple $x_{n}=(e_{i}, r_{k}, e_{j})$ using the relation code as:
\begin{equation}\label{eq:f-ijk-relcode}
  f(x_{n};\mathbf{\Theta}) = \sigma \left( \mathbf{z}^{(L)} \cdot \mathbf{v}^{(\mathrm{2})}_{j} + \mathbf{B}^{(L+1)} \cdot \mathbf{c}^{(k)}\right) .
\end{equation}

In the same way, all RMNN parameters, including $\mathbf{\Theta}=\{\mathbf{W}, \mathbf{B}, \mathbf{V}^{\mathrm{(1)}}, \mathbf{V}^{\mathrm{(2)}}, \mathbf{C}\}$, can be jointly learned based on the above maximum likelihood estimation.

The RMNN models are particularly suitable for \textit{knowledge transfer learning}, where a pre-trained model can be quickly extended to any new relation after observing a few samples from that relation.
In this case, we may estimate a new relation code based on the available new samples while keeping the whole network unchanged. Due to its small size, the new relation code can be reliably estimated from only a small number of new samples. Furthermore, model performance in all original relations will not be affected since the model and all original relation codes are not changed during transfer learning.

\section{Experiments}
\label{sec:experm}

In this section, we evaluate the proposed NAM models for various reasoning tasks.
We first describe the experimental setup and then we report the results from several reasoning tasks, including textual entailment recognition, triple classification in multi-relational KBs, commonsense reasoning and knowledge transfer learning.

\subsection{Experimental setup}

Here we first introduce some common experimental settings used for all experiments:
1) For entity or sentence representations, we represent them by composing from their word vectors as in \cite{socher2013reasoning}. All word vectors are initialized from a pre-trained skip-gram \cite{mikolov2013efficient} word embedding model, trained on a large English Wikipedia corpus.
The dimensions for all word embeddings are set to 100 for all experiments;
2) The dimensions of  all relation codes are set to 50. All relation codes are randomly initialized;
3) For network structures, we use ReLU as the nonlinear activation function and all network parameters are initialized according to  \cite{glorot2010understanding}.
Meanwhile, since the number of training examples for most probabilistic reasoning tasks is relatively small, we adopt the dropout approach \cite{hinton2012improving} during the training process to avoid the over-fitting problem;
4) During the learning process of NAMs, we need to use negative samples, which are automatically generated by  randomly perturbing positive KB triples as $\mathcal{D}^{-} = \{ (e_{i}, r_{k}, e_{\ell}) | e_{\ell} \ne e_{j} \wedge (e_{i}, r_{k}, e_{j}) \in \mathcal{D}^{+}\}$.

For each task, we use the provided development set to tune for the best training hyperparameters.
For example, we have tested the number of hidden layers among \{1, 2, 3\}, the initial learning rate  among \{0.01, 0.05, 0.1, 0.25, 0.5\}, dropout rate among \{0, 0.1, 0.2, 0.3, 0.4\}. Finally, we select the best setting based on the performance on the development set: the final model structure uses 2  hidden layers, and the learning rate and the dropout rate are set to be 0.1 and 0.2, respectively, for all the experiments.
During model training, the learning rate is halved once the performances in the development set decreases.
Both DNNs and RMNNs are trained using the stochastic gradient descend (SGD) algorithm.
We notice that the NAM models converge quickly after 30 epochs.

\subsection{Recognizing Textual Entailment}
Understanding entailment and contradiction is fundamental to language understanding.
Here we conduct experiments on a popular recognizing textual entailment (RTE) task, which aims to recognize the entailment relationship between a pair of English sentences.
In this experiment, we use the SNLI dataset in \cite{bowman2015large} to conduct 2-class RTE experiments (entailment or contradiction). All instances that are not labelled as ``entailment" are converted to contradiction in our experiments.
The SNLI dataset contains hundreds of thousands of training examples, which is useful for training a NAM model.
Since this data set does not include multi-relational data, we only investigate
the DNN structure for this task.
The final NAM result, along with the baseline performance provided in \cite{bowman2015large}, is listed in Table \ref{tab:res-RTE}.
\begin{table}[htb]
\centering
\begin{tabular}{l|c}
  \hline
  Model & Accuracy (\%) \\\hline
  Edit Distance \cite{bowman2015large} & 71.9 \\
  Classifier \cite{bowman2015large} & 72.2 \\
  Lexical Resources \cite{bowman2015large} & 75.0 \\
  \hline
  {\bf DNN} & \bf 84.7 \\
  \hline
\end{tabular}
\caption{Experimental results on the RTE task.}
\label{tab:res-RTE}
\end{table}

From the results, we can see the proposed DNN based NAM model achieves considerable improvements over various traditional methods.
This indicates that we can better model entailment relationship in natural language by
representing sentences in continuous space and conducting probabilistic reasoning with deep neural networks.

\subsection{Triple classification in multi-relational KBs}

In this section, we evaluate the proposed NAM models on two popular knowledge triple classification datasets, namely WN11 and FB13 in \cite{socher2013reasoning} (derived from WordNet and FreeBase), to predict whether some new triple relations hold based on other training facts in the database. The WN11 dataset contains 38,696 unique entities involving 11 different relations in total while the FB13 dataset covers 13 relations and 75,043 entities.
Table \ref{tab:Datasets} summarizes the statistics of these two datasets.
\begin{table}[htb]
\centering
\begin{tabular}{c|c|c|c|c|c}
  \hline
  \bf Dataset & \# R & \# Ent & \# Train & \# Dev & \# Test\\\hline
  WN11 & 11 & 38,696 & 112,581 & 2,609 & 10,544\\
  FB13 & 13 & 75,043 & 316,232 & 5,908 & 23,733\\
  \hline
\end{tabular}
\caption{The statistics for KBs triple classification datasets. \#R is the number of relations. \#Ent is the size of the entity set.}
\label{tab:Datasets}
\end{table}

The goal of knowledge triple classification is to predict whether a given triple $x_{n}=(e_{i}, r_{k}, e_{j})$ is correct or not. We first use the training data to learn NAM models. Afterwards,
we use the development set to tune a global threshold $T$ to make a binary decision: the triple is classified as true
if $f(x_{n};\mathbf{\Theta}) \ge T$; otherwise it is false.
The final accuracy is calculated based on how many triplets in the test set are classified correctly.

Experimental results on both WN11 and FB13 datasets are given in Table \ref{tab:res-WN-FB}, where we compare the two NAM models with all other methods reported on these two datasets.
The results clearly show that the NAM methods (DNNs and RMNNs) achieve comparable performance on these triple classification tasks, and both yield consistent improvement over all existing methods.
In particular, the RMNN model yields 3.7\% and 1.9\% absolute improvements over the popular neural tensor networks (NTN) \cite{socher2013reasoning} on WN11 and FB13 respectively.
Both DNN and RMNN models are much smaller than NTN in the number of parameters and they scale well as the number of relation types increases.  For example,
both DNN and RMNN models for WN11 have about 7.8 millions of parameters while NTN has about 15 millions.
Although the RESCAL and TransE models have about 4 millions of parameters for WN11, their size goes up quickly for other tasks of thousands or more relation types.
In addition, the training time of DNN and RMNN is much shorter than that of NTN or TransE since our models converge much faster. For example, we have obtained at least a 5 times speedup over NTN in WN11. 

\begin{table}[htb]
\centering
\begin{tabular}{c|c|c|c}
  \hline
  Model & WN11 & FB13 & Avg. \\\hline
  SME \cite{bordes2012joint} & 70.0 & 63.7 & 66.9 \\
  TransE \cite{bordes2013translating} & 75.9 & 81.5 & 78.7 \\
  TransH \cite{wang2014knowledge} & 78.8 & 83.3 & 81.1 \\
  TransR \cite{lin2015learning} & 85.9 & 82.5 & 84.2 \\
  NTN \cite{socher2013reasoning} & 86.2 & 90.0 & 88.1 \\
  \hline
  {\bf DNN} & \bf 89.3 & \bf 91.5 & \bf 90.4 \\
  {\bf RMNN} & \bf 89.9 & \bf 91.9 & \bf 90.9 \\
  \hline
\end{tabular}
\caption{Triple classification accuracy in WN11 and FB13.}
\label{tab:res-WN-FB}
\end{table}

\subsection{Commonsense Reasoning}

\begin{figure}[htb]
  \centering
  \includegraphics[width=7cm]{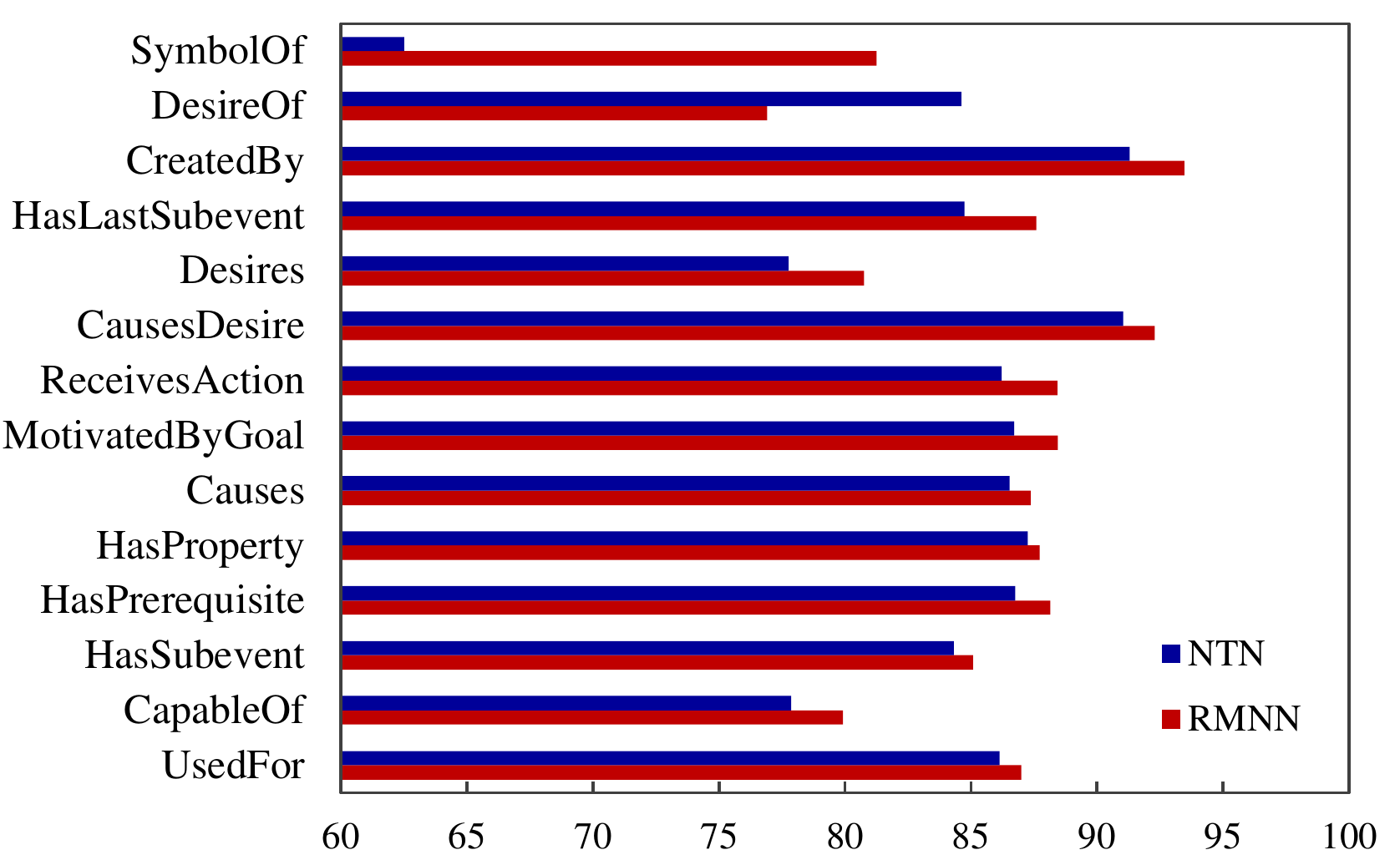}\\
  \caption{Accuracy of different relations in CN14.}
  \label{fig:CN14-rel-acc}
\end{figure}

Similar to the triple classification task \cite{socher2013reasoning}, in this work, we use the ConceptNet KB \cite{liu2004conceptnet} 
to construct a new commonsense data set, named as \textit{CN14} hereafter.
When building CN14, we first select all facts in ConceptNet related to
14 typical commonsense relations, e.g., \textit{UsedFor}, \textit{CapableOf}. (see Figure \ref{fig:CN14-rel-acc} for all 14 relations.)
Then, we randomly divide the extracted facts into three sets, Train, Dev and Test.
Finally, in order to create a test set for classification, we randomly switch entities (in the whole vocabulary) from correct triples and get a total of 2$\times$\#Test triples (half are positive samples and half are negative examples).
The statistics of CN14 are given in Table \ref{tab:Datasets-CN14}.
\begin{table}[htb]
\centering
\begin{tabular}{c|c|c|c|c|c}
  \hline
  \bf Dataset & \# R & \# Ent. & \# Train & \# Dev & \# Test\\\hline
  CN14 & 14 & 159,135 & 200,198 & 5,000 & 10,000\\
  \hline
\end{tabular}
\caption{The statistics for the CN14 dataset.}
\label{tab:Datasets-CN14}
\end{table}

The CN14 dataset is designed for answering commonsense questions like \textit{Is a camel capable of journeying across desert?}
The proposed NAM models answer this question by calculating the association probability $\Pr(E_{2}|E_{1})$ where $E_{1}= \{ \textit{camel}, \textit{capable of} \}$ and $E_{2}=\textit{journey across desert}$.
In this paper, we compare two NAM methods with the popular NTN method in \cite{socher2013reasoning} on this data set and the overall results are given in Table \ref{tab:res-CN14}.
We can see that both NAM methods outperform NTN in this task, and the DNN and RMNN models obtain similar performance.

\begin{table}[htb]
\centering
\begin{tabular}{c|c|c|c}
  \hline
  Model & \textit{Positive} & \textit{Negative} & total \\\hline 
  NTN  & 82.7 & 86.5 & 84.6 \\ \hline
  \bf{DNN} & 84.5 & 86.9 & {\bf 85.7} \\
  \bf{RMNN} & 85.1 & 87.1 & {\bf 86.1} \\ \hline
\end{tabular}
\caption{Accuracy (in \%) comparison on CN14.
}
\label{tab:res-CN14}
\end{table}

Furthermore, we show the classification accuracy of all 14  relations in CN14 for RMNN and NTN
in Figure \ref{fig:CN14-rel-acc}, which show that
the  accuracy of RMNN varies among different relations from 80.1\% (\textit{Desires}) to 93.5\% (\textit{CreatedBy}).
We notice some commonsense relations (such as \textit{Desires}, \textit{CapableOf}) are harder than the others (like \textit{CreatedBy}, \textit{CausesDesire}). RMNN overtakes NTN in almost all relations.


\subsection{Knowledge Transfer Learning}

\begin{figure}[htb]
  \centering
  \includegraphics[width=5.0cm]{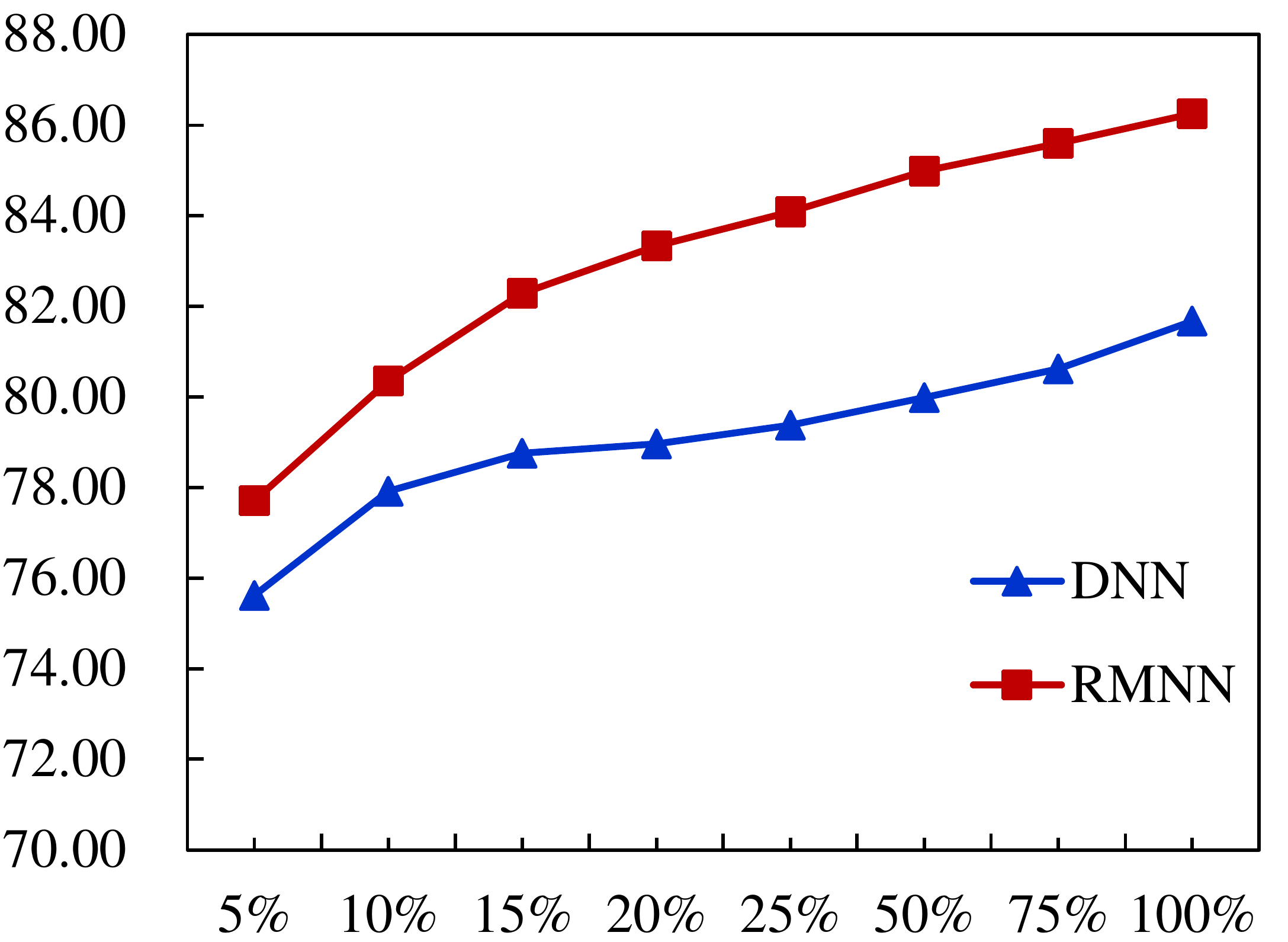}\\
  \caption{Accuracy (in \%) on the test set of a new relation \textit{CausesDesire} is shown as a function of used training samples from \textit{CausesDesire} when updating the relation code only. (Accuracy on the original relations remains as 85.7\%.)}
  \label{fig:transfer-res}
\end{figure}

Knowledge transfer between various domains is a characteristic feature and crucial cornerstone of human learning.
In this section, we evaluate the proposed NAM models for a knowledge transfer learning scenario, where
we adapt a pre-trained model to an unseen relation with only a few training samples from the new relation.
Here we randomly select a relation, e.g., \textit{CausesDesire} in CN14 for this experiment.
This relation contains only 4800 training samples and 480 test samples.
During the experiments, we use all of the other 13 relations in CN14 to train baseline NAM models (both DNN and RMNN).
During the transfer learning, we freeze all NAM parameters, including all weights and entity representations,
and only learn a new relation code for \textit{CausesDesire} from the given samples.
At last, the learned relation code (along with the original NAM models) is used to classify the new samples of \textit{CausesDesire} in the test set. Obviously, this transfer learning does not affect the model performance in the original 13 relations because the models are not changed.
Figure \ref{fig:transfer-res} shows the results of knowledge transfer learning for the relation \textit{CausesDesire} as we  increase the training samples gradually.
The result shows that RMNN performs much better than DNN in this experiment, where we can significantly improve RMNN for the new relation with only 5-20\% of the total training samples for  \textit{CausesDesire}.
This demonstrates that the structure to connect the relation code to all hidden layers leads to more effective learning of new relation codes from a relatively small number of training samples.

\begin{figure}[htb]
  \begin{minipage}[t]{0.5\linewidth}
    \centering
    \includegraphics[width=4.2cm]{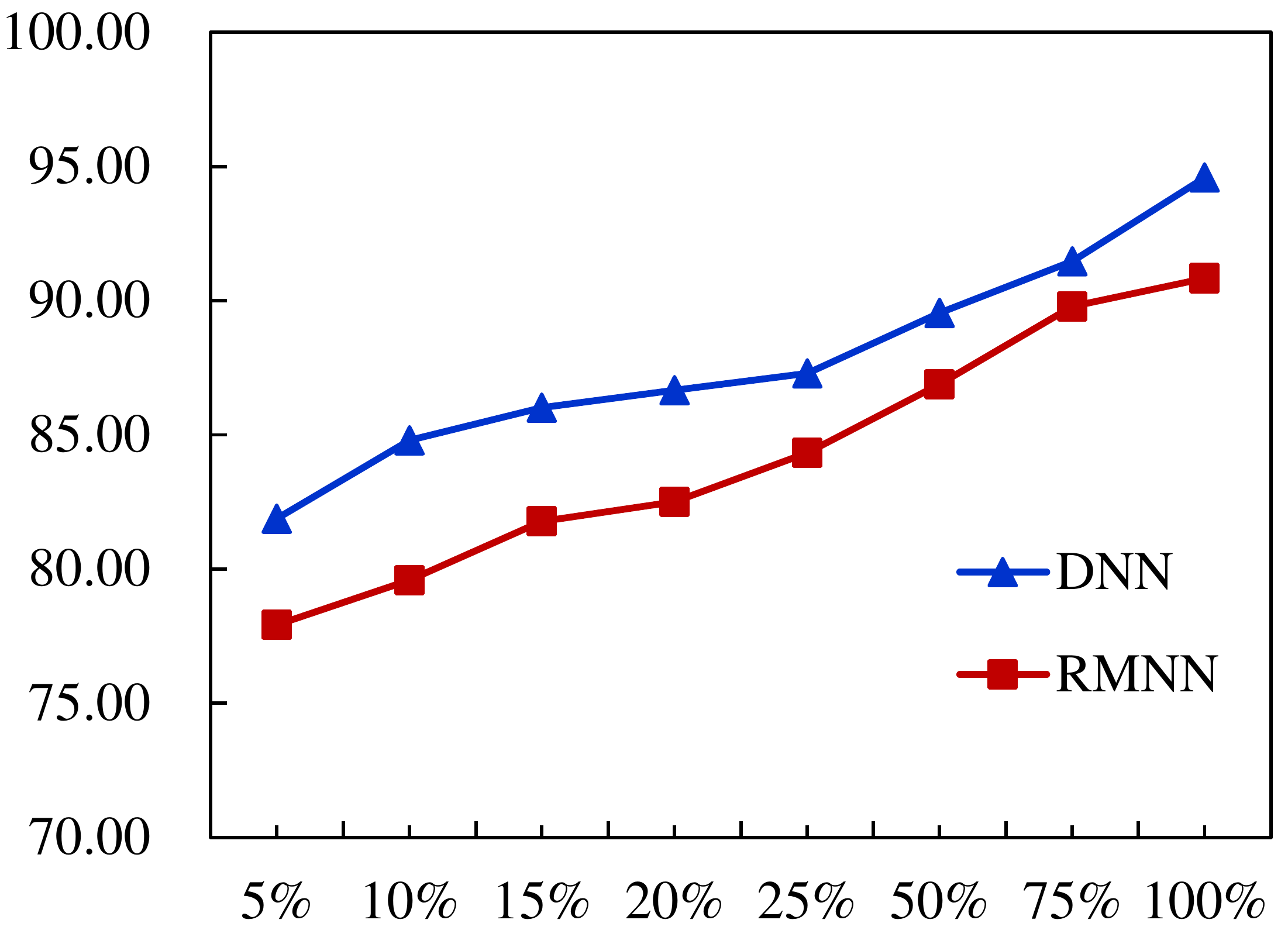}
  \end{minipage}%
  \begin{minipage}[t]{0.5\linewidth}
    \centering
    \includegraphics[width=4.2cm]{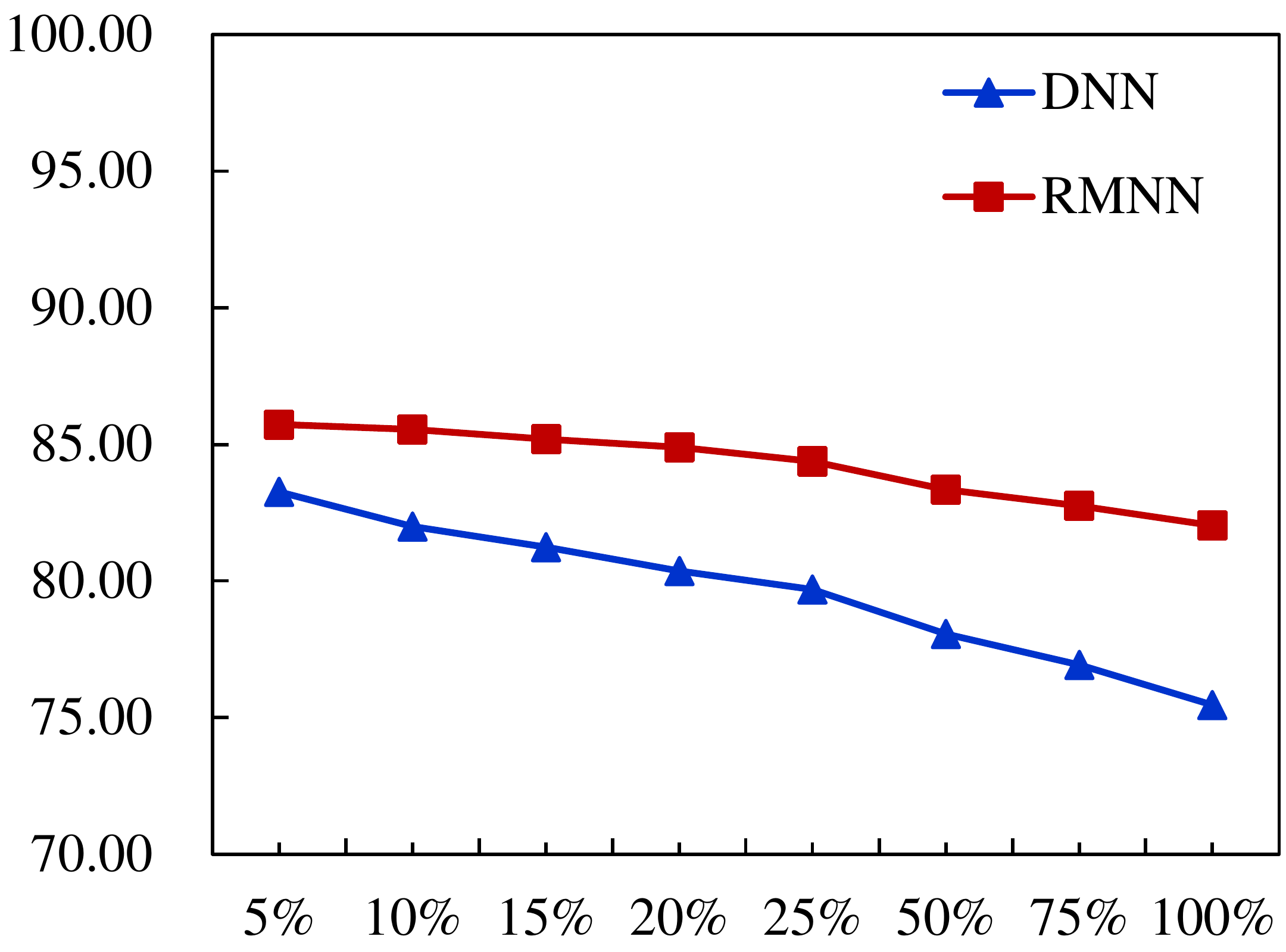}
  \end{minipage}
  \caption{Transfer learning results by updating all network parameters. The left figure shows results on the new relation while the right figure shows results on the original relations.}
  \label{fig:transfer-res-update-all}
\end{figure}

Next, we also test a more aggressive learning strategy for this transfer learning setting, where we simultaneously update all the network parameters during the learning of the new relation code. The results are shown in Figure \ref{fig:transfer-res-update-all}.
This strategy can obviously improve performance more on the new relation, especially when we add more training samples.
However, as expected, the performance on the original 13 relations deteriorates.
The DNN improves the performance on the new relation as we use all training samples (up to 94.6\%). However, the performance on the remaining 13 original relations drops dramatically from 85.6\% to 75.5\%.
Once again, RMNN shows an advantage over DNN in this transfer learning setting, where
the accuracy on the new relation increases from 77.9\% to 90.8\% but the accuracy on the original 13 relations only drop slightly from 85.9\% to 82.0\%.

\section{Extending NAMs for Winograd Schema}
\section{Data Collection}
In the previous experiments sections, all the tasks already contained manually constructed training data for us. 
However, in many cases, if we want to realize flexible commonsense reasoning under the real world conditions, obtaining the training data can also be very challenging.
More specifically, since the proposed neural association model is a typical deep learning technique, lack of training data would make it difficult for us to train a robust model.
Therefore, in this paper, we make some efforts and try to mine useful data for model training. 
As a very first step, we are now working on collecting the cause-effect relationships between a set of common words and phrases. 
We believe this type of knowledge would be a key component for modeling the association relationships between discrete events.

This section describes the idea for automatic cause-effect pair collection as well as the data collection results.
We will first introduce the common vocabulary we created for query generation. 
After that, the detailed algorithm for cause-effect pair collection will be presented. 
Finally, the following section will present the data collection results.

\subsection{Common Vocabulary and Query Generation}
To avoid the data sparsity problem, we start our work by constructing a vocabulary of very common words. 
In our current investigations, we construct a vocabulary which contains 7500 verbs and adjectives. 
As shown in Table \ref{tab:vocab}, this vocabulary includes 3000 verb words, 2000 verb phrases and 2500 adjective words.
The procedure for constructing this vocabulary is straightforward.
We first extract all words and phrases (divided by part-of-speech tags) from WordNet \cite{miller1995wordnet}.
After conducting part-of-speech tagging on a large corpus, we then get the occurrence frequencies for all those words and phrases by scanning over the tagged corpus. 
Finally, we sort those words and phrases by frequency and then select the top N results.
\begin{table}[htb]
\centering
\begin{tabular}{c|l|c}
\hline
  Set & Category & Size \\\hline
  1 & Verb words &  3000 \\
  2 & Verb phrases & 2000 \\
  3 & Adjective words & 2500 \\
\hline
\end{tabular}
\caption{Common vocabulary constructed for mining cause-effect event pairs.} 
\label{tab:vocab}
\end{table}

\subsubsection{Query Generation} 
\begin{figure}[htb]
  \centering
  \includegraphics[width=8cm]{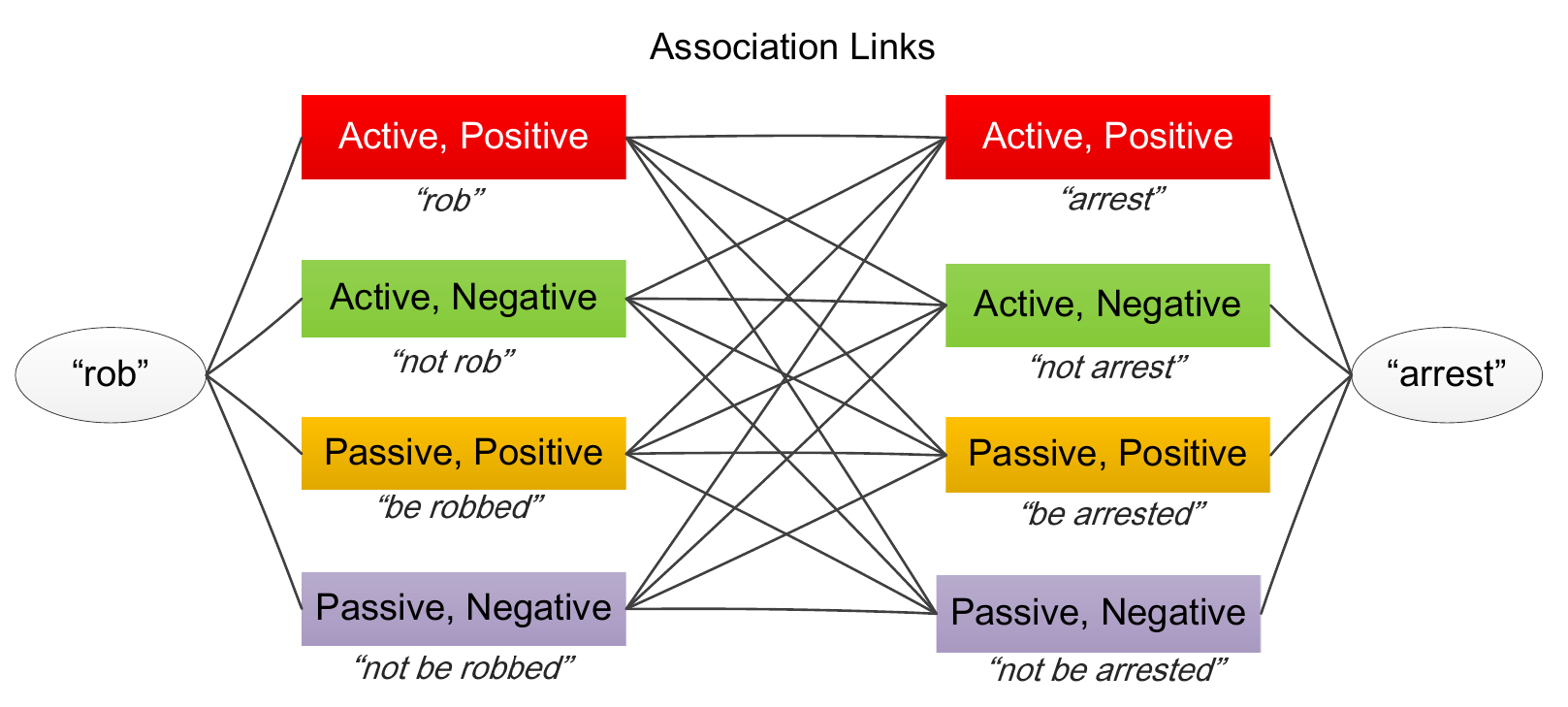}\\
  \caption{Typical 16 dimensions for a typical query.}
  \label{fig:16dim}
\end{figure}

Based on the common vocabulary, we generate search queries by pairing any two words (or phrases).
Currently we only focus on extracting the association relationships between verbs and adjectives. 
Even for this small vocabulary, the search space is very large (7.5K by 7.5K leads to tens of millions pairs).
In this work, we define several patterns for each word or phrase based on two popular semantic dimensions: 1) positive-negative, 2) active-passive \cite{osgood1952nature}. 
Using the verbs \textit{rob} and \textit{arrest} for example, each of them contains 4 patterns, i.e. (active, positive), (active, negative), (passive, positive) and (passive, negative). 
Therefore, the query formed by \textit{rob} and \textit{arrest} would contain 16 possible dimensions, as shown in Figure \ref{fig:16dim}.
The task of mining the cause-effect relationships for any two words or phrases then becomes the task of getting the number of occurrences for all the possible links.

\subsection{Automatic Cause-Effect Pair Collection}
\begin{figure}[htb]
  \centering
  \includegraphics[width=8cm]{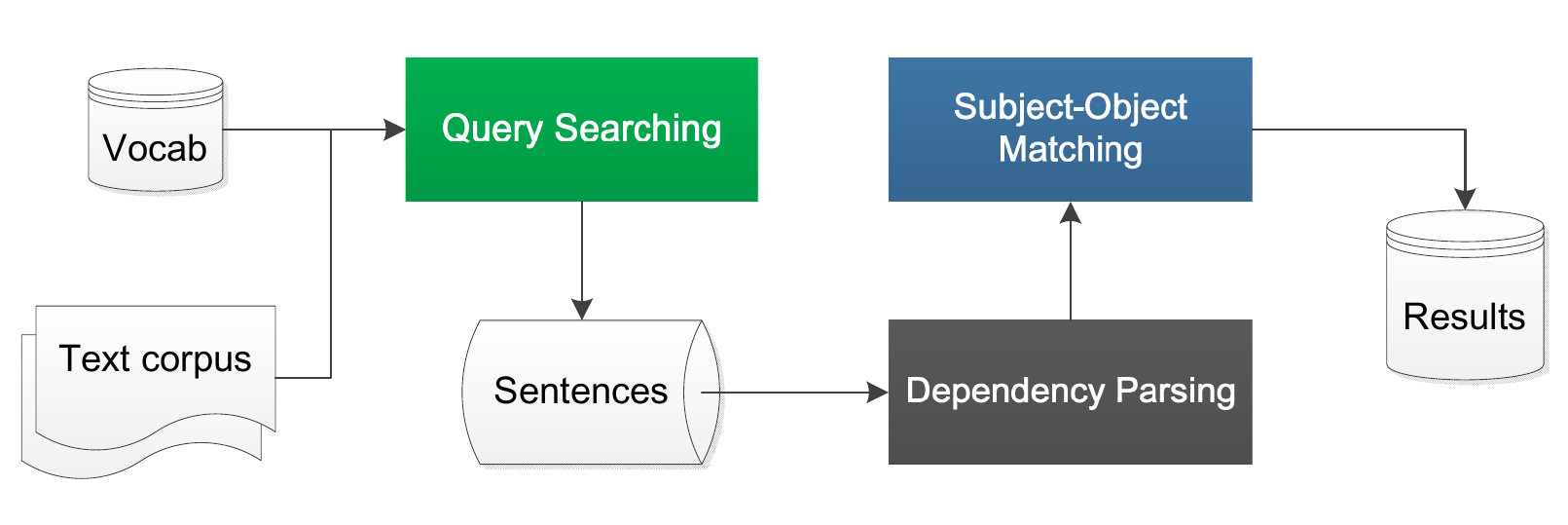}\\
  \caption{Automatic pair collection system framework.}
  \label{fig:search-system}
\end{figure}

Based on the created queries, in this section, we present the procedures for extracting cause-effect pairs from large unstructured texts. 
The overall system framework is shown in Figure \ref{fig:search-system}.

\subsubsection{Query Searching}
The goal of query searching is to find all the possible sentences that may contain the input queries. 
Since the number of queries is very large, we structure all the queries as a hashmap and conduct string matching during text scanning.
In detail, the searching program starts by conducting lemmatizing, part-of-speech tagging and dependency parsing on the source corpus. 
After it, we scan the corpus from the begining to end.
When dealing with each sentence, we will try to find the matched words (or phrases) using the hashmap. 
This strategy help us to reduce the search complexity to be linear with the size of corpus, which has been proved to be very efficient in our experiments.

\subsubsection{Subject-Object Matching}
\begin{figure}[htb]
  \centering
  \includegraphics[width=7cm]{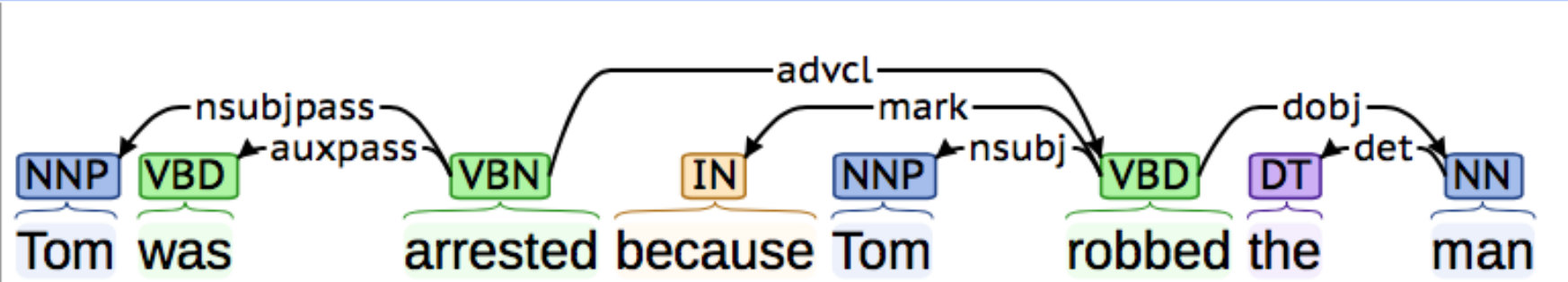}\\
  \caption{Dependency parsing result of sentence \textit{``Tom was arrested because Tom robbed the man"}.}
  \label{fig:rob-arrest}
\end{figure}

Based on the dependency parsing results, once we find one phrase of a query, we would check whether that phrase is associated with at least one subject or object in the corresponding sentence or not.
At the same time, we record whether the phrase was positive or negative, active or passive.
Moreover, for helping us to decide the cause-effect relationships, we would check whether the phrase is linked with some connective words or not. 
Typical connective words used in this work are \textit{because} and \textit{if}.
To finally extract the cause-effect pairs, we design a simple \textit{subject-object matching} rule, which is similar to the work of \cite{peng2015solving}.
1) If the two phrases in one query share the same \textit{subject}, the relationship between them is then straightforward; 
2) If the \textit{subject} of one phrase is the \textit{object} of the other phrase, then we need to apply the \textit{passive} pattern to the phrase related to the \textit{object}.
This subject-object matching idea is similar to the work proposed in \cite{peng2015solving}.
Using query (\textit{arrest}, \textit{rob}) as an example. 
Once we find sentence \textit{``Tom was arrested because Tom robbed the man"}, we obtain its dependency parsing result as shown in Figure \ref{fig:rob-arrest}.
The verb \textit{arrest} and \textit{rob} share a same subject, and the pattern for \textit{arrest} is passive, we will add the occurrence of the corresponding association link, i.e. link from the (active,positive) pattern of \textit{rob} to the (passive,positive) pattern of \textit{arrest}, by 1. 

\subsection{Data Collection Results}
Table \ref{tab:corpus} shows the corpora we used for collecting the cause-effect pairs and the corresponding data collection results.
We extract approximately 240,000 pairs from different corpora.
\begin{table}[htb]
\centering
\begin{tabular}{l|c}
\hline
  Corpus  & $\#$Result pairs \\\hline
  Gigaword \cite{graff2003english} & 117,938 \\
  Novels \cite{zhu2015aligning} &129,824 \\
  CBTest \cite{hill2015goldilocks} & 4,167 \\
  BNC \cite{burnard1995users} & 2,128\\
  \hline
\end{tabular}
\caption{Data collection results on different corpora.}
\label{tab:corpus}
\end{table}

\section{Winograd Schema Challenge}
Based on all the experiments described in the previous sections, we could conclude that the neural association model has the potential to be effective in commonsense reasoning.
To further evaluate the effectiveness of the proposed neural association model, in this paper, we conduct experiments on solving the complex \textit{Winograd Schema} challenge problems \cite{levesque2011winograd,morgenstern2016planning}.
\textit{Winograd Schema} is a commonsense reasoning task proposed in recent years, which has been treated as an alternative to the Turing Test \cite{turing1950computing}. 
This is a new AI task and it would be very interesting to see whether neural network methods are suitable for solving this problem.
This section then describes the progress we have made in attempting to meet the Winograd Schema Challenge.
For making clear what is the main task of the \textit{Winograd Schema}, we will firstly introduce it at a high level. 
Afterwards, we will introduce the system framework as well as all the corresponding modules we proposed to automatically solve the \textit{Winograd Schema} problems. 
Finally, experiments and discussions on a human annotated cause-effect dataset and discussion will be presented.

\subsection{Winograd Schema}
The \textit{Winograd Schema} (WS) evaluates a system's commonsense reasoning ability based on a traditional, very difficult natural language processing task: coreference resolution \cite{levesque2011winograd,sabawinograd}.
The \textit{Winograd Schema} problems are carefully designed to be a task that cannot be easily solved without commonsense knowledge.
In fact, even the solution of traditional coreference resolution problems relies on semantics or world knowledge \cite{rahman2011coreference,strubenon16}.
For describing the WS in detail, here we just copy some words from \cite{levesque2011winograd}.
A WS is a small reading comprehension test involving a single binary question. Here are two examples:
\begin{itemize}
\em
\item The trophy would not fit in the brown suitcase because it was too big. What was too big?
	\begin{itemize}
	\item Answer 0: the trophy
	\item Answer 1: the suitcase
	\end{itemize}
\item Joan made sure to thank Susan for all the help she had given. Who had given the help?
	\begin{itemize}
	\item Answer 0: Joan
	\item Answer 1: Susan
	\end{itemize}
\end{itemize}
The correct answers here are obvious for human beings. In each of the questions, the corresponding WS has the following four features:
\begin{enumerate}
\item Two parties are mentioned in a sentence by noun phrases. They can be two males, two females, two inanimate objects or two groups of people or objects.
\item A pronoun or possessive adjective is used in the sentence in reference to one of the parties, but is also of the right sort for the second party. In the case of males, it is ``he/him/his"; for females, it is ``she/her/her" for inanimate object it is ``it/it/its," and for groups it is ``they/them/their."
\item The question involves determining the referent of the pronoun or possessive adjective. Answer 0 is always the first party mentioned in the sentence (but repeated from the sentence for clarity), and Answer 1 is the second party.
\item There is a word (called the \textit{special} word) that appears in the sentence and possibly the question. When it is replaced by another word (called the \textit{alternate} word), everything still makes perfect sense, but the answer changes.
\end{enumerate}
Solving WS problems is not easy since the required commonsense knowledge is quite difficult to collect. 
In the following sections, we are going to describe our work on solving the \textit{Winograd Schema} problems via neural network methods.

\begin{table*}[htb]
\centering
\scriptsize
\begin{tabular}{l|c|c|c}
  \hline
  Schema texts & Verb/Adjective 1 & Verb/Adjective 2 & Verb/Adjective 3 \\\hline 
  \textit{The man couldn't lift his son because he was so weak} & \textit{weak} & \textit{not lift} & \textit{not be lifted} \\
  \textit{The man couldn't lift his son because he was so heavy} & \textit{heavy} & \textit{not lift} & \textit{not be lifted} \\
  \textit{The fish ate the worm. it was tasty} & \textit{tasty} & \textit{eat} & \textit{be eaten} \\
  \textit{The fish ate the worm. it was hungry} & \textit{hungry} & \textit{eat} & \textit{be eaten} \\
  \textit{Mary tucked her daughter Anne into bed, so that she could sleep} & \textit{tuck into bed} & \textit{be tucked into bed} & \textit{sleep} \\
  \textit{Mary tucked her daughter Anne into bed, so that she could work} & \textit{tuck into bed} & \textit{be tucked into bed} & \textit{work}\\
  \textit{Tom threw his schoolbag down to ray after he reached the top of the stairs} & \textit{reach top} & \textit{throw down} & \textit{be thrown down} \\
   \textit{Tom threw his schoolbag down to ray after he reached the bottom of the stairs} & \textit{reach bottom} & \textit{throw down} & \textit{be thrown down} \\
   \textit{Jackson was greatly influenced by Arnold, though he lived two centuries earlier} & \textit{live earlier} & \textit{influence} & \textit{be influenced} \\
   \textit{Jackson was greatly influenced by Arnold, though he lived two centuries later} & \textit{live later} & \textit{influence} & \textit{be influenced} \\
   \hline
\end{tabular}
\caption{Examples of the Cause-Effect dataset labelled from the Winograd Schema Challenge.}
\label{tab:ce-winograd-data}
\end{table*}

\subsection{System Framework}
In this paper, we propose that the commonsense knowledge required in many \textit{Winograd Schema} problems could be formulized as some association relationships between discrete events.
Using sentence ``\textit{Joan made sure to thank Susan for all the help she had given}" as an example, the commonsense knowledge is that the man who receives help should thank to the man who gives help to him. 
We believe that by modeling the association between event \textit{receive help} and \textit{thank}, \textit{give help} and \textit{thank}, we can make the decision by comparing the association probability \textit{$\Pr(\textrm{thank}|\textrm{receive help})$} and \textit{$\Pr(\textrm{thank}|\textrm{give help})$}.
If the models are well trained, we should get the inequality \textit{$\Pr(\textrm{thank}|\textrm{receive help}) > \Pr(\textrm{thank}|\textrm{give help})$}.
Following this idea, we propose to utilize the data constructed from the previous section and extend the NAM models for solving WS problems.
Here we design two frameworks for training NAM models.
\begin{itemize}
\item \textit{TransMat}-NAM:  We design to apply four linear transformation matrices, i.e., matrices of (active, positive), (active, negative), (passive, positive) and (passive, negative), for transforming both the cause event and the effect event. After it, we then use NAM for model the cause-effect association relationship between any cause and effect events.
   \begin{figure}[htb]
      \centering
      \includegraphics[width=8cm]{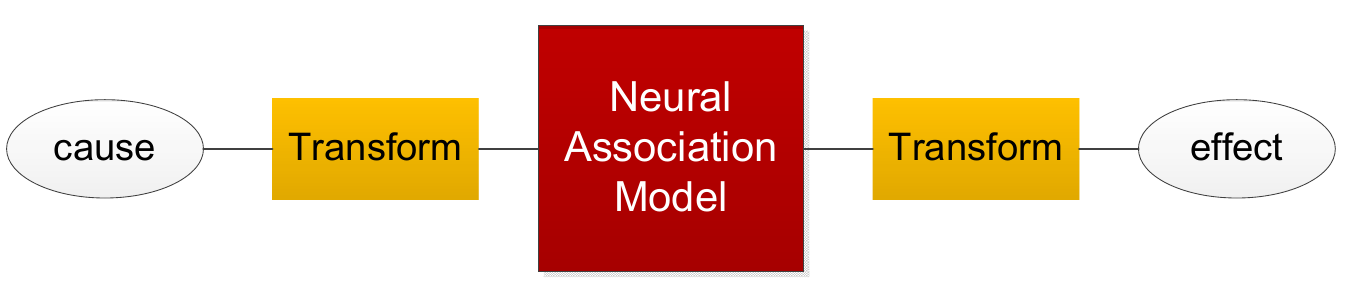}\\
      \caption{The model framework for \textit{TransMat}-NAM.}
    \end{figure}
\item \textit{RelationVec}-NAM: On the other hand, in this configuration, we treat all the typical 16 dimensions shown in Figure \ref{fig:16dim} as distinct relations. So there are 16 relation vectors in the corresponding NAM models. Currently we use the RMNN structure for NAM.
      \begin{figure}[htb]
      \centering
      \includegraphics[width=6.75cm]{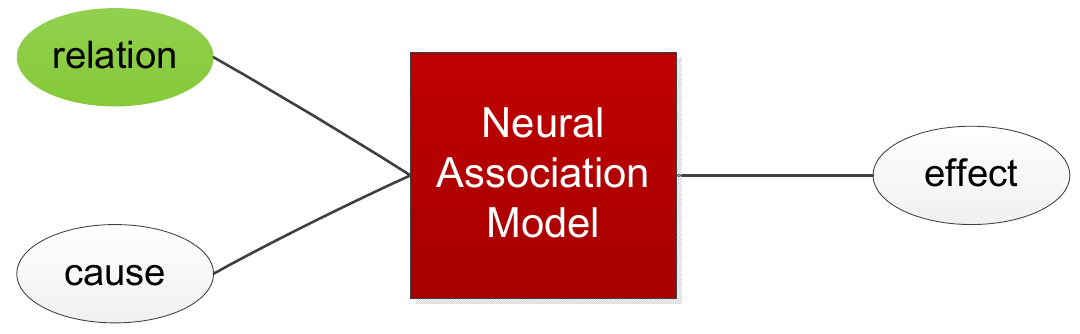}\\
      \caption{The model framework for \textit{RelationVec}-NAM.}
    \end{figure}
\end{itemize}

Training the NAM models based on these two configurations is straightforward.
All the network parameters, including the relation vectors and the linear transformation matrices, are learned by the standard stochastic gradient descend algorithm.

\subsection{Experiments}
In this section, we will introduce our current experiments on solving the \textit{Winograd Schema} problems.
We will first select a cause-effect dataset constructed from the standard WS dataset.
Subsequently, experimental setup will be described in detail.
After presenting the experimental results, discussions would be made at the end of this section.
\subsubsection{Cause-Effect Dataset Labelling} 
In this paper, based on the WS dataset available at \url{http://www.cs.nyu.edu/faculty/davise/papers/WinogradSchemas/WS.html}, we labelled 78 cause-effect problems among all 278 available WS questions for our experiments.
Table \ref{tab:ce-winograd-data} shows some typical examples.
For each WS problem, we label three verb (or adjective) phrases for the corresponding two parities and the pronoun.
In the labelled phrases, we also record the corresponding patterns for each word respectively.
Using word \textit{lift} for example, we will generate \textit{lift} for its active and positive pattern, \textit{not lift} for its active and negative pattern, \textit{be lifted} for its passive and positive pattern, and \textit{not be lifted} for its passive and negative pattern.
For example, in sentence ``\textit{The man couldn't lift his son because he was so weak}", we identify \textit{weak}, \textit{not lift} and \textit{not be lifted} for \textit{he}, \textit{the man} and \textit{son} resspectively.
The commonsense is that somebody who is \textit{weak} would more likely to has the effect \textit{not lift} rather than \textit{not be lifted}.
The main work of NAM for solving this problem is to calculate the association probability between these phrases. 

\subsubsection{Experimental setup}
The setup for NAM on this cause-effect task is similar to the settings on the previous tasks.
For representing the phrases in neural association models, we use the bag-of-word (BOW) approach for composing phrases from pre-trained word vectors.
Since the vocabulary we use in this experiment contains only 7500 common verbs and adjectives, there are some out-of-vocabulary (OOV) words in some phrases.
Based on the BOW method, a phrase would be useless if all the words it contains are OOV.
In this paper, we remove all the testing samples with useless phrases which results in 70 testing cause-effect samples.
For network settings, we set the embedding size to 50 and the dimension of relation vectors to 50. 
We set 2 hidden layers for the NAM models and all the hidden layer sizes are set to 100.
The learning rate is set to 0.01 for all the experiments. 
At the same time, to better control the model training, we set the learning rates for learning all the embedding matrices and the relation vectors to 0.025.

Negative sampling is very important for model training for this task.
In the \textit{TransMat}-NAM system, we generate negative samples by randomly selecting different patterns with respect to the pattern of the \textit{effect} event in the positive samples.
For example, if the positive training sample is ``\textit{hungry (active, positive)} causes \textit{eat (active, positive)}", we may generate negative samples like ``\textit{hungry (active, positive)} causes \textit{eat (passive, positive)}", or ``\textit{hungry (active, positive)} causes \textit{eat (active, negative)}".
In the \textit{RelationVec}-NAM system, the negative sampling method is much more straightforward, i.e., we will randomly select a different \textit{effect} event from the whole vocabulary.
In the example shown here, the possible negative sample would be ``\textit{hungry (active, positive)} causes \textit{happy (active, positive)}", or ``\textit{hungry (active, positive)} causes \textit{talk (active, positive)}" and so on.

\subsubsection{Results}
The experimental results are shown in Table \ref{tab:res-winograd-ce}.
From the results, we find that the proposed NAM models achieve about 60\% accuracy on the cause-effect dataset constructed from \textit{Winograd Schemas}.
More specifically, the \textit{RelationVec}-NAM system performs slightly better than the \textit{TransMat}-NAM system.
\begin{table}[htb]
\centering
\begin{tabular}{l|c}
  \hline
  Model & Accuracy (\%) \\\hline 
  \textit{TransMat}-NAM & \bf 58.6 (41 / 70)\\
  \textit{RelationVec}-NAM & \bf 61.4 (43 / 70)\\
  \hline
\end{tabular}
\caption{Results of NAMs on the \textit{Winograd Schema} Cause-Effect dataset.}
\label{tab:res-winograd-ce}
\end{table}

In the testing results, we find the NAM performs well on some testing examples. 
For instance, in the \textit{call phone} scenario, the proposed NAM generates the corresponding association probabilities as follows.
\begin{itemize}
\em 
\item Paul tried to call George on the phone, but he wasn't successful. Who was not successful?
	\begin{itemize}
	\item Paul: $\Pr(\text{not successful} | \text{call})$ = \rm 0.7299
	\item \em George: $\Pr(\text{not successful} | \text{be called})$ = \rm 0.5430
	\item Answer: \em Paul
	\end{itemize}
\item Paul tried to call George on the phone, but he wasn't available. Who was not available?
	\begin{itemize}
	\item Paul: $\Pr(\text{not available} | \text{call})$ = \rm 0.6859
	\item \em George: $\Pr(\text{not available} | \text{be called})$ = \rm 0.8306
	\item Answer: \em George
	\end{itemize}
\end{itemize}

For these testing examples, we find our model can answer the questions by correctly calculating the association probabilities.
The probability \textit{$\Pr(\text{not successful} | \text{call})$} is larger than \textit{$\Pr(\text{not successful} | \text{be called})$} while the probability \textit{$\Pr(\text{not available} | \text{call})$} is smaller than \textit{$\Pr(\text{not available} | \text{be called})$}.
Those simple inequality relationships between the association probabilities are very reasonable in our commonsense.
Here are some more examples:

\begin{itemize}
\em 
\item Jim yelled at Kevin because he was so upset. Who was upset?
	\begin{itemize}
	\item \em Jim: $\Pr(\text{yell} | \text{be upset})$ = \rm 0.9296
	\item \em Kevin: $\Pr(\text{be yelled} | \text{be upset})$ = \rm 0.8785
	\item Answer: \em Jim
	\end{itemize}
\item Jim comforted Kevin because he was so upset. Who was upset?
	\begin{itemize} \em
	\item \em Jim: $\Pr(\text{comfort} | \text{be upset})$ = \rm 0.0282
	\item \em Kevin: $\Pr(\text{be comforted} | \text{be upset})$ = \rm 0.5657
	\item Answer: \em Kevin
	\end{itemize}
\end{itemize}

This example also conveys some commonsense knowledge in our daily life.
We all know that somebody who is upset would be more likely to yell at other people.
Meanwhile, it is also more likely that they would be be comforted by other people.

\section{Conclusions}
\label{sec:final}
In this paper, we have proposed neural association models (NAM) for probabilistic reasoning.
We use neural networks to model association probabilities between any two events in a domain.
In this work, we have investigated two model structures, namely DNN and RMNN, for NAMs.
Experimental results on several reasoning tasks have shown that both DNNs and RMNNs can outperform the existing methods.
This paper also reports some preliminary results to use NAMs for knowledge transfer learning.
We have found that the proposed RMNN model can be quickly adapted to a new relation without sacrificing the performance in the original relations.
After proving the effectiveness of the NAM models, we apply it to solve more complex commonsense reasoning problems, i.e., the \textit{Winograd Schemas} \cite{levesque2011winograd}.
To support model training in this task, we propose a straightforward method to collect associative phrase pairs from text corpora.
Experiments conducted on a set of \textit{Winograd Schema} problems have indicated the neural association model does solve some problems successfully.
However, it is still a long way to finally achieving automatic commonsense reasoning.

\section*{Acknowledgments}
We want to thank Prof. Gary Marcus of New York University for his useful comments on commonsense reasoning.
We also want to thank Prof. Ernest Davis, Dr. Leora Morgenstern and Dr. Charles Ortiz for their wonderful organizations for making the first Winograd Schema Challenge happen.
This paper was supported in part by the Science and Technology Development of Anhui Province, China (Grants No. 2014z02006), the Fundamental Research Funds for the Central Universities (Grant No. WK2350000001) and the Strategic Priority Research
Program of the Chinese Academy of Sciences (Grant No. XDB02070006).

\bibliographystyle{aaai}
\bibliography{nam_aaai16}

\begin{thebibliography}{}

\bibitem[\protect\citeauthoryear{Bengio \bgroup et al\mbox.\egroup
  }{2003}]{bengio2003neural}
Bengio, Y.; Ducharme, R.; Vincent, P.; and Janvin, C.
\newblock 2003.
\newblock A neural probabilistic language model.
\newblock {\em The Journal of Machine Learning Research} 3:1137--1155.

\bibitem[\protect\citeauthoryear{Bordes \bgroup et al\mbox.\egroup
  }{2012}]{bordes2012joint}
Bordes, A.; Glorot, X.; Weston, J.; and Bengio, Y.
\newblock 2012.
\newblock Joint learning of words and meaning representations for open-text
  semantic parsing.
\newblock In {\em Proceedings of AISTATS},  127--135.

\bibitem[\protect\citeauthoryear{Bordes \bgroup et al\mbox.\egroup
  }{2013}]{bordes2013translating}
Bordes, A.; Usunier, N.; Garcia-Duran, A.; Weston, J.; and Yakhnenko, O.
\newblock 2013.
\newblock Translating embeddings for modeling multi-relational data.
\newblock In {\em Proceedings of NIPS},  2787--2795.

\bibitem[\protect\citeauthoryear{Bowman \bgroup et al\mbox.\egroup
  }{2015}]{bowman2015large}
Bowman, S.~R.; Angeli, G.; Potts, C.; and Manning, C.~D.
\newblock 2015.
\newblock A large annotated corpus for learning natural language inference.
\newblock {\em arXiv preprint arXiv:1508.05326}.

\bibitem[\protect\citeauthoryear{Bowman}{2013}]{bowman2013can}
Bowman, S.~R.
\newblock 2013.
\newblock Can recursive neural tensor networks learn logical reasoning?
\newblock {\em arXiv preprint arXiv:1312.6192}.

\bibitem[\protect\citeauthoryear{Burnard}{1995}]{burnard1995users}
Burnard, L.
\newblock 1995.
\newblock Users reference guide british national corpus version 1.0.

\bibitem[\protect\citeauthoryear{Collobert \bgroup et al\mbox.\egroup
  }{2011}]{collobert2011natural}
Collobert, R.; Weston, J.; Bottou, L.; Karlen, M.; Kavukcuoglu, K.; and Kuksa,
  P.
\newblock 2011.
\newblock Natural language processing (almost) from scratch.
\newblock {\em The Journal of Machine Learning Research} 12:2493--2537.

\bibitem[\protect\citeauthoryear{Getoor}{2007}]{getoor2007introduction}
Getoor, L.
\newblock 2007.
\newblock {\em Introduction to statistical relational learning}.
\newblock MIT press.

\bibitem[\protect\citeauthoryear{Glorot and
  Bengio}{2010}]{glorot2010understanding}
Glorot, X., and Bengio, Y.
\newblock 2010.
\newblock Understanding the difficulty of training deep feedforward neural
  networks.
\newblock In {\em Proceedings of AISTATS},  249--256.

\bibitem[\protect\citeauthoryear{Graff \bgroup et al\mbox.\egroup
  }{2003}]{graff2003english}
Graff, D.; Kong, J.; Chen, K.; and Maeda, K.
\newblock 2003.
\newblock English gigaword.
\newblock {\em Linguistic Data Consortium, Philadelphia}.

\bibitem[\protect\citeauthoryear{Hill \bgroup et al\mbox.\egroup
  }{2015}]{hill2015goldilocks}
Hill, F.; Bordes, A.; Chopra, S.; and Weston, J.
\newblock 2015.
\newblock The goldilocks principle: Reading children's books with explicit
  memory representations.
\newblock {\em arXiv preprint arXiv:1511.02301}.

\bibitem[\protect\citeauthoryear{Hinton \bgroup et al\mbox.\egroup
  }{2012}]{hinton2012improving}
Hinton, G.~E.; Srivastava, N.; Krizhevsky, A.; Sutskever, I.; and
  Salakhutdinov, R.~R.
\newblock 2012.
\newblock Improving neural networks by preventing co-adaptation of feature
  detectors.
\newblock {\em arXiv preprint arXiv:1207.0580}.

\bibitem[\protect\citeauthoryear{Hornik, Stinchcombe, and
  White}{1990}]{hornik1990universal}
Hornik, K.; Stinchcombe, M.; and White, H.
\newblock 1990.
\newblock Universal approximation of an unknown mapping and its derivatives
  using multilayer feedforward networks.
\newblock {\em Neural networks} 3(5):551--560.

\bibitem[\protect\citeauthoryear{Jensen}{1996}]{jensen1996introduction}
Jensen, F.~V.
\newblock 1996.
\newblock {\em An introduction to Bayesian networks}, volume 210.
\newblock UCL press London.

\bibitem[\protect\citeauthoryear{Koller and
  Friedman}{2009}]{koller2009probabilistic}
Koller, D., and Friedman, N.
\newblock 2009.
\newblock {\em Probabilistic graphical models: principles and techniques}.
\newblock MIT press.

\bibitem[\protect\citeauthoryear{LeCun, Bengio, and
  Hinton}{2015}]{lecun2015deep}
LeCun, Y.; Bengio, Y.; and Hinton, G.
\newblock 2015.
\newblock Deep learning.
\newblock {\em Nature} 521(7553):436--444.

\bibitem[\protect\citeauthoryear{Levesque, Davis, and
  Morgenstern}{2011}]{levesque2011winograd}
Levesque, H.~J.; Davis, E.; and Morgenstern, L.
\newblock 2011.
\newblock The winograd schema challenge.
\newblock In {\em AAAI Spring Symposium: Logical Formalizations of Commonsense
  Reasoning}.

\bibitem[\protect\citeauthoryear{Lin \bgroup et al\mbox.\egroup
  }{2015}]{lin2015learning}
Lin, Y.; Liu, Z.; Sun, M.; Liu, Y.; and Zhu, X.
\newblock 2015.
\newblock Learning entity and relation embeddings for knowledge graph
  completion.
\newblock In {\em Proceedings of AAAI}.

\bibitem[\protect\citeauthoryear{Liu and Singh}{2004}]{liu2004conceptnet}
Liu, H., and Singh, P.
\newblock 2004.
\newblock Conceptnet: a practical commonsense reasoning toolkit.
\newblock {\em BT technology journal} 22(4):211--226.

\bibitem[\protect\citeauthoryear{McCarthy}{1986}]{mccarthy1986applications}
McCarthy, J.
\newblock 1986.
\newblock Applications of circumscription to formalizing common-sense
  knowledge.
\newblock {\em Artificial Intelligence} 28(1):89--116.

\bibitem[\protect\citeauthoryear{Mikolov \bgroup et al\mbox.\egroup
  }{2013}]{mikolov2013efficient}
Mikolov, T.; Chen, K.; Corrado, G.; and Dean, J.
\newblock 2013.
\newblock Efficient estimation of word representations in vector space.
\newblock {\em arXiv preprint arXiv:1301.3781}.

\bibitem[\protect\citeauthoryear{Miller}{1995}]{miller1995wordnet}
Miller, G.~A.
\newblock 1995.
\newblock Wordnet: a lexical database for english.
\newblock {\em Communications of the ACM} 38(11):39--41.

\bibitem[\protect\citeauthoryear{Minsky}{1988}]{minsky1988society}
Minsky, M.
\newblock 1988.
\newblock {\em Society of mind}.
\newblock Simon and Schuster.

\bibitem[\protect\citeauthoryear{Morgenstern, Davis, and
  Ortiz~Jr}{2016}]{morgenstern2016planning}
Morgenstern, L.; Davis, E.; and Ortiz~Jr, C.~L.
\newblock 2016.
\newblock Planning, executing, and evaluating the winograd schema challenge.
\newblock {\em AI Magazine} 37(1):50--54.

\bibitem[\protect\citeauthoryear{Mueller}{2014}]{mueller2014commonsense}
Mueller, E.~T.
\newblock 2014.
\newblock {\em Commonsense Reasoning: An Event Calculus Based Approach}.
\newblock Morgan Kaufmann.

\bibitem[\protect\citeauthoryear{Nair and Hinton}{2010}]{nair2010rectified}
Nair, V., and Hinton, G.~E.
\newblock 2010.
\newblock Rectified linear units improve restricted boltzmann machines.
\newblock In {\em Proceedings of ICML},  807--814.

\bibitem[\protect\citeauthoryear{Neapolitan}{2012}]{neapolitan2012probabilistic}
Neapolitan, R.~E.
\newblock 2012.
\newblock {\em Probabilistic reasoning in expert systems: theory and
  algorithms}.
\newblock CreateSpace Independent Publishing Platform.

\bibitem[\protect\citeauthoryear{Nickel \bgroup et al\mbox.\egroup
  }{2015}]{nickel2015review}
Nickel, M.; Murphy, K.; Tresp, V.; and Gabrilovich, E.
\newblock 2015.
\newblock A review of relational machine learning for knowledge graphs.
\newblock {\em arXiv preprint arXiv:1503.00759}.

\bibitem[\protect\citeauthoryear{Nickel, Tresp, and
  Kriegel}{2012}]{nickel2012factorizing}
Nickel, M.; Tresp, V.; and Kriegel, H.-P.
\newblock 2012.
\newblock Factorizing {YAGO}: scalable machine learning for linked data.
\newblock In {\em Proceedings of WWW},  271--280.
\newblock ACM.

\bibitem[\protect\citeauthoryear{Osgood}{1952}]{osgood1952nature}
Osgood, C.~E.
\newblock 1952.
\newblock The nature and measurement of meaning.
\newblock {\em Psychological bulletin} 49(3):197.

\bibitem[\protect\citeauthoryear{Pearl}{1988}]{pearl1988probabilistic}
Pearl, J.
\newblock 1988.
\newblock Probabilistic reasoning in intelligent systems: Networks of plausible
  reasoning.

\bibitem[\protect\citeauthoryear{Peng, Khashabi, and
  Roth}{2015}]{peng2015solving}
Peng, H.; Khashabi, D.; and Roth, D.
\newblock 2015.
\newblock Solving hard coreference problems.
\newblock {\em Urbana} 51:61801.

\bibitem[\protect\citeauthoryear{Rahman and Ng}{2011}]{rahman2011coreference}
Rahman, A., and Ng, V.
\newblock 2011.
\newblock Coreference resolution with world knowledge.
\newblock In {\em Proceedings of the 49th Annual Meeting of the Association for
  Computational Linguistics: Human Language Technologies-Volume 1},  814--824.
\newblock Association for Computational Linguistics.

\bibitem[\protect\citeauthoryear{Richardson and
  Domingos}{2006}]{richardson2006markov}
Richardson, M., and Domingos, P.
\newblock 2006.
\newblock Markov logic networks.
\newblock {\em Machine learning} 62(1-2):107--136.

\bibitem[\protect\citeauthoryear{Saba}{2015}]{sabawinograd}
Saba, W.
\newblock 2015.
\newblock On the winograd schema challenge.

\bibitem[\protect\citeauthoryear{Socher \bgroup et al\mbox.\egroup
  }{2013}]{socher2013reasoning}
Socher, R.; Chen, D.; Manning, C.~D.; and Ng, A.
\newblock 2013.
\newblock Reasoning with neural tensor networks for knowledge base completion.
\newblock In {\em Proceedings of NIPS},  926--934.

\bibitem[\protect\citeauthoryear{Strube}{2016}]{strubenon16}
Strube, M.
\newblock 2016.
\newblock The (non) utility of semantics for coreference resolution (corbon
  remix).
\newblock In {\em NAACL 2016 workshop on Coreference Resolution Beyond
  OntoNotes}.

\bibitem[\protect\citeauthoryear{Turing}{1950}]{turing1950computing}
Turing, A.~M.
\newblock 1950.
\newblock Computing machinery and intelligence.
\newblock {\em Mind} 59(236):433--460.

\bibitem[\protect\citeauthoryear{Wang \bgroup et al\mbox.\egroup
  }{2014}]{wang2014knowledge}
Wang, Z.; Zhang, J.; Feng, J.; and Chen, Z.
\newblock 2014.
\newblock Knowledge graph embedding by translating on hyperplanes.
\newblock In {\em Proceedings of AAAI},  1112--1119.
\newblock Citeseer.

\bibitem[\protect\citeauthoryear{Xue \bgroup et al\mbox.\egroup
  }{2014}]{xue2014fast}
Xue, S.; Abdel-Hamid, O.; Jiang, H.; Dai, L.; and Liu, Q.
\newblock 2014.
\newblock Fast adaptation of deep neural network based on discriminant codes
  for speech recognition.
\newblock {\em Audio, Speech, and Language Processing, IEEE/ACM Trans. on}
  22(12):1713--1725.

\bibitem[\protect\citeauthoryear{Zhu \bgroup et al\mbox.\egroup
  }{2015}]{zhu2015aligning}
Zhu, Y.; Kiros, R.; Zemel, R.; Salakhutdinov, R.; Urtasun, R.; Torralba, A.;
  and Fidler, S.
\newblock 2015.
\newblock Aligning books and movies: Towards story-like visual explanations by
  watching movies and reading books.
\newblock In {\em Proceedings of the IEEE International Conference on Computer
  Vision},  19--27.

\end{thebibliography}

\end{document}